\documentclass[11pt]{article}

\newif\iffullbuild
\fullbuildtrue

\usepackage[final]{acl}

\usepackage{times}
\usepackage{latexsym}

\usepackage[T1]{fontenc}

\usepackage[utf8]{inputenc}

\usepackage{microtype}

\usepackage{inconsolata}

\usepackage{graphicx}

\usepackage{svg}
\usepackage{color}
\usepackage{tabularx}
\usepackage{tabu}
\usepackage{booktabs}
\usepackage{makecell}
\usepackage{float}

\usepackage{hyperref}
\usepackage{tablefootnote}
\usepackage{xspace}

\usepackage{algorithm}
\usepackage{algpseudocode}
\usepackage{amsmath}
\usepackage{sansmath}
\usepackage{amssymb}
\usepackage{amsthm}
\usepackage{marvosym}
\usepackage{xcolor}
\usepackage{colortbl}
\usepackage{caption}
\usepackage{tcolorbox}
\usepackage{subcaption}
\usepackage{pdfpages}
\usepackage{arydshln}
\usepackage[normalem]{ulem}
\usepackage{multirow}
\usepackage{cleveref}
\usepackage{pifont}
\definecolor{rowgray}{gray}{0.95}
\definecolor{ZJUBlue}{RGB}{0,63,136}
\definecolor{YaleBlue}{HTML}{00356B}
\definecolor{TongjiBlue}{HTML}{005BAC}
\definecolor{UCASBlue}{HTML}{0072BC}

\newcommand{\ZJUZ}{\textcolor{ZJUBlue}{\textsf{\textbf{\textit{Z}}}}}
\newcommand{\YaleY}{\textcolor{YaleBlue}{\textsf{\textbf{\textit{Y}}}}}
\newcommand{\TongjiT}{\textcolor{TongjiBlue}{\textsf{\textbf{\textit{T}}}}}
\newcommand{\UCASU}{\textcolor{UCASBlue}{\textsf{\textbf{\textit{U}}}}}
\newcommand\blfootnote[1]{%
  \begingroup
  \renewcommand\thefootnote{}\footnotetext{#1}%
  \endgroup
}

\usepackage{listings}
\usepackage{enumitem}
\usepackage{amssymb}
\tcbuselibrary{skins, breakable, listings}

\makeatletter
\g@addto@macro\normalsize{%
  \setlength{\abovedisplayskip}{6pt}%
  \setlength{\belowdisplayskip}{6pt}%
  \setlength{\abovedisplayshortskip}{6pt}%
  \setlength{\belowdisplayshortskip}{6pt}%
}
\makeatother
\normalsize

\newcommand{\ours}{\textsc{AnTrap}\xspace}
\newcommand{\modelnum}{16\xspace}
\newcommand{\orgnum}{7\xspace}
\newcommand{\tasknum}{236\xspace}

\newcommand{\gmark}{\textsuperscript{$\dagger$}}

\newcommand{\githuburl}{\url{https://github.com/gguogan/AnTrap}}

\title{Are Android GUI Agents Robust Against Runtime Anomalies?\\ \ours: Evaluating Agents in Dynamic Adversarial Environments}

\author{
  \textbf{Guo Gan\textsuperscript{\ZJUZ}} \quad
  \textbf{Yilun Zhao\textsuperscript{\YaleY}} \quad
  \textbf{Cong Chen\textsuperscript{\ZJUZ}} \quad
  \textbf{Jinbiao Wei\textsuperscript{\YaleY}} \quad
  \textbf{Tingyu Song\textsuperscript{\UCASU}} \quad\\
  \textbf{Zheyuan Yang\textsuperscript{\TongjiT}} \quad
  \textbf{Lin Fu\textsuperscript{\ZJUZ}} \quad
  \textbf{Hong Zhou\textsuperscript{\ZJUZ\Letter}}
\\[0.8em]
  \textsuperscript{\ZJUZ}Zhejiang University \quad\quad
  \textsuperscript{\YaleY}Yale University \quad\quad
  \textsuperscript{\TongjiT}Tongji University \quad\quad \\
  \textsuperscript{\UCASU}University of Chinese Academy of Sciences
}

\begin{document}
\maketitle
\blfootnote{\textsuperscript{\Letter}Corresponding author: Hong Zhou \href{mailto:zhouhong_zju@zju.edu.cn}{<zhouhong\_zju@zju.edu.cn>}. Our code will be available at \githuburl.}
\begin{abstract}
GUI agents often encounter dynamic anomalies when deployed on Android devices, from unexpected pop-ups to action misuse, yet existing benchmarks lack systematic evaluation of agent robustness against runtime anomalies.
We introduce \ours, a comprehensive benchmark that injects dynamic perturbations into agent execution trajectories. We propose a taxonomy organizing real-world anomalies into four layers (State, Thinking, Action and Round) with ten fine-grained subcategories, and develop a construction pipeline that preserves task solvability while introducing realistic adversarial conditions.
Evaluating \modelnum leading GUI models, we reveal universal vulnerability to dynamic anomalies, with even the strongest models suffering significant performance degradation.
Furthermore, we conduct GRPO training in both original and adversarial environments to validate our benchmark, separating environment-learnable anomalies from reasoning-bottlenecked ones.
Our findings show that while single-step traps at state and action layers are largely addressable through adversarial reinforcement learning, deep contextual traps, like state deadlock, expose intrinsic limitations that cannot be resolved by training in environments with traps alone.

\end{abstract}

\section{Introduction}
 \begin{figure*}[!ht]
    \centering
    \includegraphics[width=\textwidth]{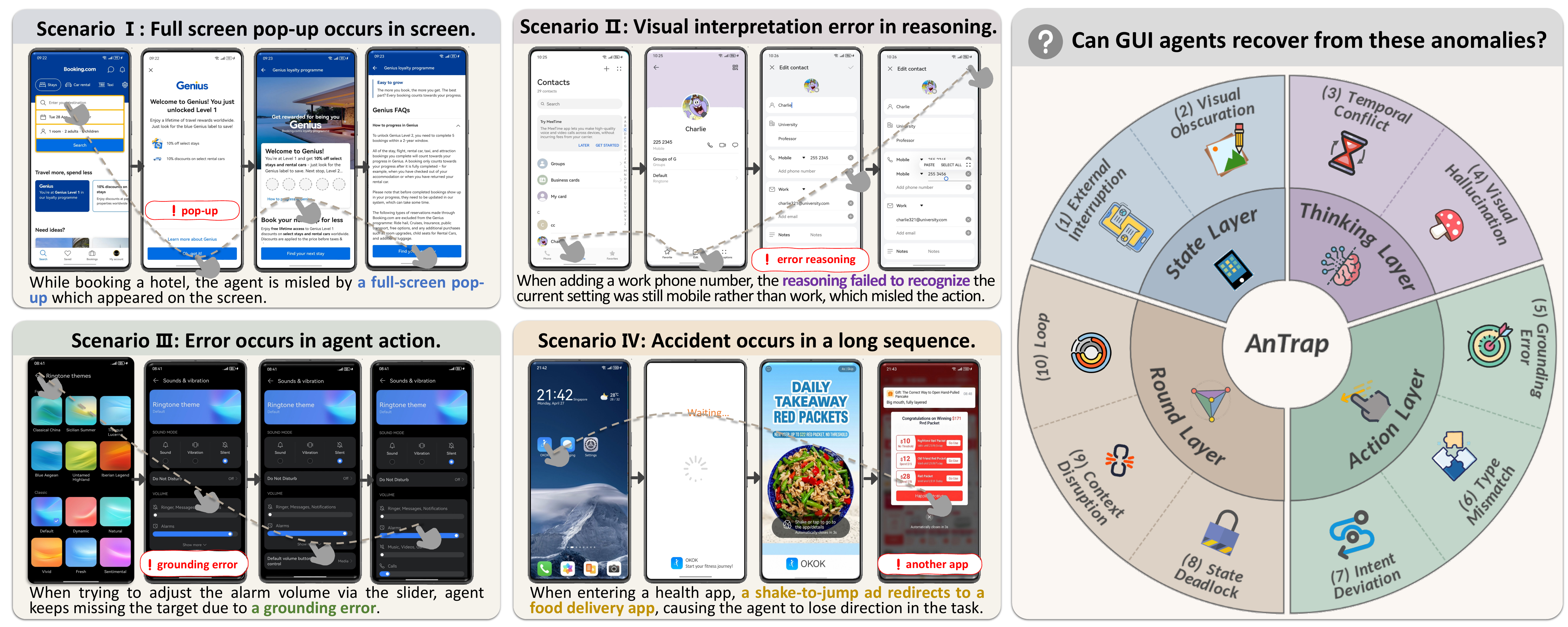}
    \vspace{-2em}
    \caption{Four scenarios of runtime anomalies in Android agent tasks from our pilot study, including pop-ups, reasoning errors, grounding errors, and long-sequence accidents. \ours taxonomy organizes such anomalies into four layers (State, Thinking, Action, Round) and ten subcategories.}
    \label{fig:motivation}
\end{figure*}

Android agents aim to accomplish complex device operation tasks given instructions from human users~\cite{bai2024digirl, chen2025gui, wang2025ui, gan-etal-2026-android}. Recently-released agentic models
have shown remarkable long-sequence planning and execution performance on dynamic benchmarks, including AndroidLab~\cite{xu2024androidlab}, AndroidWorld~\cite{rawles2024androidworld}, and MobileWorld~\cite{kong2025mobileworldbenchmarkingautonomousmobile}.

Despite these contributions, the robustness of agents against Android runtime anomalies remains largely unexplored.
This gap is critical since GUI agents frequently encounter runtime anomalies from two sources that require recovery in real-world mobile execution.
First, agents face external environment perturbations such as application ad pop-ups, system glitches and rendering delays that often disrupt correct task execution~\cite{xu2026smanbench, zhang2025attackingvisionlanguagecomputeragents, xu2026mobilebenchv2realisticcomprehensivebenchmark}.
Second, agents make internal errors themselves during operation, such as hallucinating GUI elements in reasoning, grounding inaccurately or misusing actions, yet they consistently struggle to recover independently~\cite{yuan2025agent, wu2025gui}. Consequently, the ability to recover from both types of anomalies is essential for the real-world deployment of Android agents~\cite{zhang2026dont, zhai2026guide, wu2025backtrackagentenhancingguiagent}. However, existing benchmarks either use predefined adversarial static trajectories for single-step testing~\cite{xu2026smanbench, wei2026step}, or lack systematic anomaly injection in dynamic settings, thus failing to cover the full spectrum of real-world runtime challenges we discussed~\cite{sun2026ambibenchbenchmarkingmobilegui, gong2026venusbenchmobilechallengingusercentricbenchmark}.

To bridge this gap, we propose \ours, a novel benchmark suite designed with challenging dynamic \textsc{\underline{An}}droid \textsc{\underline{Trap}}s to assess agent recovery capabilities against external environmental perturbations and internal errors.
We construct an original set of 236 tasks within a dynamic environment extended from AndroidWorld~\cite{rawles2024androidworld}. To systematically test the robustness of agents against runtime anomalies, we conduct a pilot study in real-world scenarios and introduce a comprehensive taxonomy for granular adversarial evaluation on Android. As detailed in Figure~\ref{fig:motivation}, it consists of four primary layers containing \textit{state}, \textit{thinking}, \textit{action}, and \textit{round} based on the execution loop principle, along with ten specific subcategories for constructing adversarial tasks.

On \ours, we conduct extensive experiments, evaluating \modelnum agentic models from \orgnum organizations known for their leading performance in GUI tasks.
Our experiment results reveal that current agents generally exhibit performance drops when facing adversarial perturbations under dynamic traps. We attribute this to the agents' inadequate integration of historical trajectory context, and a scarcity of adversarial training environments and data.

Consequently, to analyze the unique challenges presented by \ours, we conduct online Group Relative Policy Optimization (GRPO) in both the original and adversarial environments. The resulting performance separates environment-learnable challenges from reasoning-bottlenecked ones: state and action perturbations are largely recovered through adversarial training, whereas multi-step contextual traps such as execution loops resist improvement even under dedicated training.

The contributions are summarized as follows:
\begin{itemize}[leftmargin=*, noitemsep, topsep=-0.4em]
    \item We introduce a comprehensive taxonomy of runtime anomalies for Android GUI agents, categorizing real-world deployment challenges into four layers and ten subcategories.
    \item We design the \ours suite, which includes an online environment and a benchmark with dynamic traps to evaluate agent performance under runtime anomalies.
    \item We analyze current agents using \ours, revealing their vulnerability to runtime anomalies.
    \item We conduct GRPO experiments that provide insights for future agent development by distinguishing between environment-learnable challenges and reasoning-bottlenecked issues.
\end{itemize}

\section{Related Work}
\definecolor{rowbeige}{HTML}{F6F3EC}
\definecolor{rowbeigedark}{HTML}{E5DDC4}
\newcommand{\cmark}{\textcolor{green!55!black}{$\checkmark$}}
\newcommand{\xmark}{\textcolor{red!75!black}{\ding{55}}}

\begin{table*}[!ht]
\centering
\small
\renewcommand{\arraystretch}{1.05}
\renewcommand\tabularxcolumn[1]{m{#1}}
\begin{tabularx}{\textwidth}{l c c c X}
\toprule
\textbf{Benchmark} & \textbf{Online Env} & \textbf{Stochastic Tasks} & \textbf{Intervention} & \textbf{Task Target} \\ \midrule

\multicolumn{5}{c}{\textbf{\textit{Computer Agentic Systems}}} \\ \hdashline
\rowcolor{rowbeige} ScreenSpot Pro \cite{li2025screenspotpro} & \xmark & $-$ & N/A & Grounding \\
\rowcolor{rowbeige} WebArena \cite{zhou2024webarenarealisticwebenvironment} & \cmark & \xmark & N/A & General Execution \\
\rowcolor{rowbeige} OSWorld \cite{xie2024osworldbenchmarkingmultimodalagents} & \cmark & \xmark & N/A & General Execution \\
\rowcolor{rowbeige} It's a TRAP! \cite{korgul2025itstraptaskredirectingagent} & \cmark & \xmark & Dynamic & Safety \\
\rowcolor{rowbeige} AutoElicit \cite{jones2026benigninputsleadsevere} & \cmark & \xmark & Dynamic & Safety \\
\rowcolor{rowbeige} LPS Bench \cite{chen2026lpsbenchbenchmarkingsafetyawareness} & \cmark & \xmark & Dynamic & Safety \\
\rowcolor{rowbeige} Robustness ScreenSpot \cite{zhao2025robustness} & \xmark & $-$ & Static & Robustness \\ \midrule

\multicolumn{5}{c}{\textbf{\textit{Mobile Agentic Systems}}} \\ \hdashline
\rowcolor{rowbeige} ScreenSpot V2\cite{wu2024osatlasfoundationactionmodel} & \xmark & $-$ & N/A & Grounding \\
\rowcolor{rowbeige} AITW \cite{rawles2023androidwildlargescaledataset} & \xmark & $-$ & N/A & General Execution \\
\rowcolor{rowbeige} AndroidControl \cite{li2024effects} & \xmark & $-$ & N/A & General Execution \\
\rowcolor{rowbeige} GUI Odyssey \cite{lu2025guiodyssey} & \xmark & $-$ & N/A & General Execution \\
\rowcolor{rowbeige} AndroidLab \cite{xu2024androidlab} & \cmark & \xmark & N/A & General Execution \\
\rowcolor{rowbeige} AndroidWorld \cite{rawles2024androidworld} & \cmark & \cmark & N/A & General Execution \\
\rowcolor{rowbeige} MobileWorld \cite{kong2025mobileworldbenchmarkingautonomousmobile} & \cmark & \xmark & N/A & General Execution \\
\rowcolor{rowbeige} MobileBench-OL \cite{wu2026mobilebencholcomprehensivechinesebenchmark} & \cmark & \xmark & Dynamic & General Execution \\
\rowcolor{rowbeige} AmbiBench \cite{sun2026ambibenchbenchmarkingmobilegui} & \cmark & \xmark & Dynamic & Ambiguity \\
\rowcolor{rowbeige} SMAN Bench \cite{xu2026smanbench} & \xmark & $-$ & Static & Robustness \\
\hdashline
\noalign{\vskip 2pt}
\rowcolor{rowbeigedark} \textbf{\ours (Ours)} & \textbf{\cmark} & \textbf{\cmark} & \textbf{Dynamic} & \textbf{Robustness} \\ \bottomrule
\end{tabularx}
\caption{Comparison of recent graphical user interface benchmarks. Unlike prior evaluation suites that predominantly focus on general execution or passive robustness, \textbf{\ours} injects runtime anomalies during execution to assess agent robustness in mobile environments with stochastic instructions.}
\label{tab:benchmark_comparison}
\end{table*}

\subsection{Benchmarks for Mobile Usage}
\label{sec:related_1}
Benchmarks for GUI agents generally fall into static and dynamic categories as shown in Table~\ref{tab:benchmark_comparison}.
Static benchmarks~\cite{rawles2023androidwildlargescaledataset,zhang2024androidzoochainofactionthoughtgui,li2024effects,lu2025guiodyssey} focus on single-step operations and grounding accuracy, neglecting the capacity of agents for long-sequence tool utilization within dynamic environments.
Dynamic benchmarks including AndroidLab~\cite{xu2024androidlab}, AndroidWorld~\cite{rawles2024androidworld}, and MobileWorld~\cite{kong2025mobileworldbenchmarkingautonomousmobile} construct interactive frameworks with foundational Android emulators, which assess agent performance through continuous task execution within a live operating system~\cite{wu2026mobilebencholcomprehensivechinesebenchmark}.
However, these benchmarks primarily evaluate the overall performance of mobile agents under stable and clean conditions while overlooking their robustness.
Some works have acknowledged this issue, but they either focus solely on visual noise~\cite{zhao2025robustnessguigroundingmodels, xu2026smanbench} or the ambiguity of instructions~\cite{sun2026ambibenchbenchmarkingmobilegui}, lacking a systematic evaluation of agents' recovery capabilities when facing runtime anomalies.
To address this, \ours introduces a granular taxonomy of runtime anomalies which cover both external environment perturbations and agent internal error cases.

\subsection{Agent Robustness to Anomalies}
Across diverse domains, evaluating and improving the robustness of agents against anomalies has become an important research direction.
Literatures on computer usage~\cite{jones2026benigninputsleadsevere, zhao2025robustness,wei2026opencomputer} evaluate agent robustness against anomalies in desktop environments across diverse dimensions, such as visual~\cite{chen2025evaluating}, semantic~\cite{korgul2025itstraptaskredirectingagent}, and privacy aspects~\cite{chen2026lpsbenchbenchmarkingsafetyawareness}.
In coding domains, RoTBench~\cite{ye2024rotbench} evaluates the robustness of agents against dynamic perturbations, particularly focusing on noisy execution environments and tool interfaces.
However, in mobile scenarios, relevant benchmarks~\cite{xu2026smanbench, sun2026ambibenchbenchmarkingmobilegui, gong2026venusbenchmobilechallengingusercentricbenchmark, wu2026mobilebencholcomprehensivechinesebenchmark} are either static or dynamic but only focus on limited anomalies as we mention in Section~\ref{sec:related_1}.
Several methods focus on robust execution~\cite{zhang2026dont}. Methods like Agent-R~\cite{yuan2025agent} enhance general agents by pre-collecting error recovery trajectories for supervised fine-tuning, while the mobile-scenario method GUI-Reflection~\cite{wu2025gui} focuses on improving agent recovery by identifying errors, undoing actions, and getting back on track. However, these methods only capture random errors and accidents, lacking controllable environments and systematic analysis across different anomaly types.
Our work fills this gap by injecting fine-grained exogenous environmental and endogenous cognitive perturbations in a dynamic manner, specifically designed to evaluate the recovery capabilities of agents.

 \begin{figure*}[!ht]
    \centering
    \includegraphics[width=\textwidth]{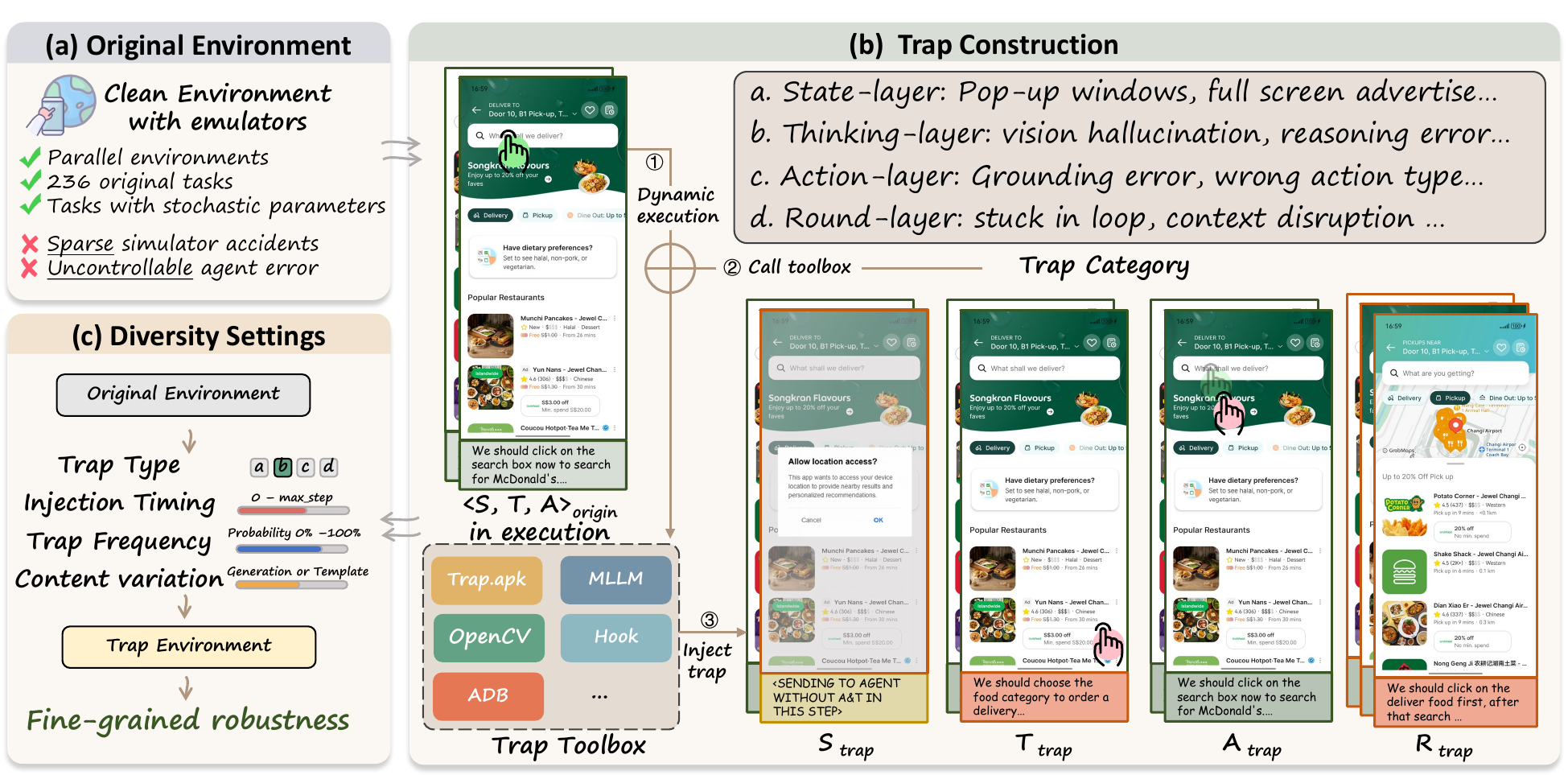}
    \vspace{-2em}
    \caption{Overview of \ours. (a) Original Environment: a base emulator that supports parallel evaluation over base tasks and environments. (b) \ours Construction: during dynamic execution, we intercept the agent's $\langle S, T, A \rangle$ loop and inject four-layer anomalies through the trap toolbox. (c) Diversity Settings: trap type, injection timing, frequency, and content source can be flexibly configured to create diverse and realistic dynamic traps.}
    \label{fig:pipeline}
\end{figure*}

\section{Android Trap}
In this section, we first present the \textbf{preliminary} for benchmark design (Section~\ref{sec:preliminary}). Second, we introduce \textbf{taxonomy} of dynamic traps (Section~\ref{sec:taxonomy}). Then, we present \textbf{\ours construction}  including original task curation and adversarial traps injection (Section~\ref{sec:antrap_construction}). Finally, we provide \textbf{data analysis} including benchmark statistics and human validation for the reliability of \ours. (Section~\ref{sec:data_release}).

\subsection{Preliminary}
\label{sec:preliminary}
\paragraph{Task Formulation.}
We propose \ours, a benchmark designed to assess the robustness of GUI agents against runtime anomalies.
A conventional GUI task expects an agent to fulfill a user instruction via an execution loop of multiple interaction rounds $R=(S, T, A)$, comprising the environment-provided state $S$, the agent's internal thinking $T$, and the executed action $A$. However, real-world execution frequently involves dynamic anomalies including environmental disturbances and agents' own errors.
Formally, to simulate these anomalies, we inject dynamic perturbations $L_{\text{trap}}$ into the execution loop at step $t$, converting clean environments into adversarial ones as follows:\vspace{-0.3em}
\begin{equation}
X \xrightarrow{L_{\text{trap}}} \tilde{X},  \text{where } X \in \{S_t, T_t, A_t, R_{t-k:t}\},
\label{eq:trap_op}
\end{equation}
where $\tilde{X}$ denotes the perturbed counterpart of $X$, $k$ is the window size of consecutive steps.

\paragraph{Prerequisite and Desiderata.}
Evaluating robustness against runtime anomalies introduces two challenges not addressed by existing benchmarks~\cite{xu2026smanbench}. First, the evaluation suite requires \textit{fine-grained, category-level diversity} to reflect a broad range of real-world anomalies which agents encounter upon deployment, rather than relying on sparse anomalies incidentally triggered within standard dynamic environments. Second, constructing these adversarial conditions requires a \textit{systematic and scalable} methodology so that each injected perturbation remains controllable and aligned with the intended evaluation signal.
To realize these, we introduce a taxonomy of runtime anomalies that trap GUI agents. Subsequently, we detail an injection pipeline that systematically constructs scenarios with traps for each taxonomy category to ensure expansive task diversity.

\subsection{Taxonomy of Dynamic Traps}
\label{sec:taxonomy}
Through a pilot analysis of agents' trajectories collected from real-device deployment tasks as described in Appendix~\ref{app:pilot_work}, our taxonomy \textit{STAR} consists of four layer-based categories: \textit{State}, \textit{Thinking}, \textit{Action}, and \textit{Round} layers based on the GUI agent execution loop. As shown in Figure~\ref{fig:motivation} and Table~\ref{tab:trap_taxonomy_short}: (1) \textit{State Layer $S_{\text{trap}}$} focuses on one-step anomalies at the state level, including unexpected pop-up windows, rendering errors, and transient UI glitches; (2) \textit{Thinking Layer $T_{\text{trap}}$} focuses on errors at the thinking level, such as temporal misalignments and visual hallucinations; (3) \textit{Action Layer $A_{\text{trap}}$} focuses on issues in action selection, including grounding errors, action type errors, and inconsistencies between actions and thinking; (4) \textit{Round Layer $R_{\text{trap}}$} indicates errors spanning several steps, like long-time deadlocks, system interruptions, and falling into loops. These four categories are further divided into ten subcategories.

\begin{table}[ht]
\centering
\small
\renewcommand{\arraystretch}{1.10}
\scalebox{0.78}{
\begin{tabular}{lp{6cm}}
\toprule[.1em]
 \textbf{Subcategory} & \textbf{Definition} \\
\midrule
\multicolumn{2}{c}{\textbf{Category 1: State Layer}} \\
\noalign{\vskip 0.5ex}
(1) External Interruption & Unexpected windows, system alerts or application jumps. \\
(2) Visual Obscuration & Rendering errors or image loading failures obscuring critical interface. \\
\midrule
\multicolumn{2}{c}{\textbf{Category 2: Thinking Layer}} \\
\noalign{\vskip 0.5ex}
(3) Temporal Conflict & False or outdated observations that lead the agent to reason about a misaligned state. \\
(4) Visual Hallucination & Hallucinate text, icons, buttons, relationships, or details that do not exist on the screenshot. \\
\midrule
\multicolumn{2}{c}{\textbf{Category 3: Action Layer}} \\
\noalign{\vskip 0.5ex}
(5) Grounding Error & Executed action coordinates deviating from the intended target UI components. \\
(6) Type Mismatch & Misuse of action types, including confusing click, long press, and double-click; or calling undefined tools in action space. \\
(7) Intent Deviation & Executed actions diverging from the generated internal reasoning and planning logic. \\
\midrule
\multicolumn{2}{c}{\textbf{Category 4: Round Layer}} \\
\noalign{\vskip 0.5ex}
(8) State Deadlock & An unresponsive interface where environment state updates stall completely despite actions. \\
(9) Context Disruption & Sudden desktop jumps or unrelated application switches breaking execution. \\
(10) Loop & Loop in a repeated state cycle.\\
\bottomrule[.1em]
\end{tabular}
}
\caption{Systematic taxonomy of dynamic traps in the \ours benchmark. Comprehensive programmatic implementation details are provided in the Appendix~\ref{app:implementation}.}
\label{tab:trap_taxonomy_short}
\end{table}

\subsection{\ours Construction}
\label{sec:antrap_construction}
\paragraph{Original Task Curation.}
We use AndroidWorld~\cite{rawles2024androidworld} as our base clean environment due to its dynamic nature and extensive usage. To enhance the diversity of the baseline tasks, we manually augment the original suite of 116 tasks by introducing new scenarios within this environment, yielding a total of \tasknum tasks with stochastic variations, all of which undergo meticulous manual checking to ensure their operational feasibility and technical soundness. Specific details are provided in the Appendix~\ref{app:implementation}.
\begin{table*}[t]
\centering
\footnotesize
\setlength{\tabcolsep}{4pt}
\renewcommand{\arraystretch}{1.25}
\definecolor{tabgreen}{HTML}{AFCE9C}
\definecolor{tabred}{HTML}{E0AC9B}

\definecolor{rowgray1}{HTML}{DEE0E5}
\definecolor{rowgray2}{HTML}{E3DFE1}
\definecolor{rowgray3}{HTML}{DFE3DE}
\definecolor{rowgray4}{HTML}{E5DFD9}

\definecolor{rowgray}{HTML}{E8E8E8}
\definecolor{rowbeige}{HTML}{F6F3EC}
\resizebox{\textwidth}{!}{%
\renewcommand{\arraystretch}{1.25}
\begin{tabular}{l c p{4pt} cc p{4pt} cc p{4pt} ccc p{4pt} ccc p{4pt} c}
\toprule
\multirow{2}{*}{\textbf{Model}} & \multirow{2}{*}{\textbf{Orig}} & & \multicolumn{2}{c}{\textbf{S-Layer}} & & \multicolumn{2}{c}{\textbf{T-Layer}} & & \multicolumn{3}{c}{\textbf{A-Layer}} & & \multicolumn{3}{c}{\textbf{R-Layer}} & & \multirow{2}{*}{\textbf{Avg}} \\
\cmidrule(lr){4-5} \cmidrule(lr){7-8} \cmidrule(lr){10-12} \cmidrule(lr){14-16}
 &  &  & \textbf{Ext.I.} & \textbf{Vis.O.} &  & \textbf{Tmp.C.} & \textbf{Vis.H.} &  & \textbf{Grd.E.} & \textbf{Typ.M.} & \textbf{Int.D.} &  & \textbf{St.DL.} & \textbf{Ctx.D.} & \textbf{Loop} &  &  \\
\midrule
\rowcolor{rowgray1}\multicolumn{18}{l}{\textit{Human Baseline}} \\
Human Annotator & 94.1 &  & 94.1 & \cellcolor{tabred!3} 92.4 &  & - & - &  & 94.1 & 93.6 & - &  & 93.2 & \cellcolor{tabred!2} 92.8 & - &  & 93.4 \\
\midrule
\rowcolor{rowgray2}\multicolumn{18}{l}{\textit{Proprietary Models}} \\
Claude-Sonnet-4.6 & 74.2 &  & \cellcolor{tabred!20} 64.8 & \cellcolor{tabred!17} 66.1 &  & \cellcolor{tabred!16} 66.9 & \cellcolor{tabred!4} 72.0 &  & \cellcolor{tabred!9} 69.9 & \cellcolor{tabred!14} 67.8 & \cellcolor{tabred!10} 69.5 &  & \cellcolor{tabred!24} 63.1 & \cellcolor{tabred!24} 63.1 & \cellcolor{tabred!27} 61.9 &  & \cellcolor{tabred!16} 66.5 \\
Gemini-3-Pro & 72.9 &  & \cellcolor{tabred!24} 61.9 & \cellcolor{tabred!9} 68.6 &  & \cellcolor{tabred!27} 60.2 & \cellcolor{tabred!22} 62.7 &  & \cellcolor{tabred!9} 68.6 & \cellcolor{tabred!27} 60.6 & \cellcolor{tabred!14} 66.1 &  & \cellcolor{tabred!20} 63.6 & \cellcolor{tabred!25} 61.4 & \cellcolor{tabred!23} 62.3 &  & \cellcolor{tabred!20} 63.6 \\
GPT-5.4 & 65.3 &  & \cellcolor{tabred!22} 55.1 & \cellcolor{tabred!25} 53.8 &  & \cellcolor{tabred!16} 57.6 & 64.8 &  & - & \cellcolor{tabred!7} 61.9 & \cellcolor{tabred!7} 61.9 &  & \cellcolor{tabred!33} 50.0 & \cellcolor{tabred!28} 52.5 & \cellcolor{tabred!29} 51.7 &  & \cellcolor{tabred!19} 56.6 \\
GPT-5.4-Mini & 42.8 &  & \cellcolor{tabred!25} 31.4 & \cellcolor{tabred!39} 25.0 &  & \cellcolor{tabred!5} 40.3 & \cellcolor{tabred!28} 29.7 &  & - & \cellcolor{tabred!29} 29.2 & \cellcolor{tabred!21} 33.1 &  & \cellcolor{tabred!24} 31.8 & \cellcolor{tabred!17} 34.7 & \cellcolor{tabred!7} 39.4 &  & \cellcolor{tabred!21} 32.7 \\
\midrule
\rowcolor{rowgray3}\multicolumn{18}{l}{\textit{Open-Source Instruct Models}} \\
GUI-Owl-1.5-32B-Instruct & 68.2 &  & \cellcolor{tabred!22} 58.1 & \cellcolor{tabred!12} 62.7 &  & \cellcolor{tabred!24} 57.2 & \cellcolor{tabred!6} 65.3 &  & \cellcolor{tabred!11} 63.1 & \cellcolor{tabred!9} 64.0 & \cellcolor{tabred!4} 66.1 &  & \cellcolor{tabred!10} 63.6 & \cellcolor{tabred!12} 62.7 & \cellcolor{tabred!15} 61.0 &  & \cellcolor{tabred!12} 62.4 \\
GUI-Owl-1.5-8B-Instruct & 64.8 &  & \cellcolor{tabred!21} 55.1 & \cellcolor{tabgreen!9} 66.9 &  & \cellcolor{tabred!14} 58.1 & 65.3 &  & 65.7 & \cellcolor{tabred!3} 63.1 & \cellcolor{tabred!10} 60.2 &  & \cellcolor{tabred!21} 55.1 & \cellcolor{tabred!16} 57.2 & \cellcolor{tabred!13} 58.5 &  & \cellcolor{tabred!9} 60.5 \\
Qwen3-VL-4B-Instruct & 54.7 &  & \cellcolor{tabred!37} 37.7 & \cellcolor{tabred!5} 52.1 &  & \cellcolor{tabred!14} 47.9 & \cellcolor{tabred!51} 31.4 &  & \cellcolor{tabred!8} 50.8 & \cellcolor{tabred!8} 50.8 & \cellcolor{tabred!12} 49.2 &  & \cellcolor{tabred!26} 42.8 & \cellcolor{tabred!25} 43.2 & \cellcolor{tabred!5} 52.1 &  & \cellcolor{tabred!18} 45.8 \\
Qwen3-VL-8B-Instruct & 52.1 &  & \cellcolor{tabred!30} 38.1 & \cellcolor{tabgreen!7} 53.8 &  & \cellcolor{tabred!23} 41.5 & \cellcolor{tabred!11} 47.0 &  & \cellcolor{tabred!5} 49.6 & \cellcolor{tabred!4} 50.0 & \cellcolor{tabred!8} 48.3 &  & \cellcolor{tabred!23} 41.5 & \cellcolor{tabred!14} 45.3 & \cellcolor{tabred!9} 47.9 &  & \cellcolor{tabred!12} 46.3 \\
\midrule
\rowcolor{rowgray4}\multicolumn{18}{l}{\textit{Open-Source Thinking Models}} \\
GUI-Owl-1.5-32B-Think & 69.5 &  & \cellcolor{tabred!20} 60.2 & \cellcolor{tabred!3} 67.8 &  & \cellcolor{tabred!37} 52.5 & \cellcolor{tabred!4} 67.4 &  & \cellcolor{tabred!4} 67.4 & \cellcolor{tabred!8} 65.7 & \cellcolor{tabred!9} 65.3 &  & \cellcolor{tabred!30} 55.5 & \cellcolor{tabred!33} 54.2 & \cellcolor{tabred!2} 68.2 &  & \cellcolor{tabred!15} 62.4 \\
GUI-Owl-1.5-8B-Think & 66.1 &  & \cellcolor{tabred!28} 53.0 & 66.5 &  & \cellcolor{tabred!16} 58.5 & \cellcolor{tabred!12} 60.2 &  & \cellcolor{tabgreen!5} 67.4 & \cellcolor{tabred!8} 62.3 & \cellcolor{tabred!5} 63.6 &  & \cellcolor{tabred!6} 63.1 & \cellcolor{tabred!19} 57.2 & 65.3 &  & \cellcolor{tabred!9} 61.7 \\
GUI-Owl-7B & 63.1 &  & \cellcolor{tabred!43} 43.2 & \cellcolor{tabred!12} 57.6 &  & \cellcolor{tabred!47} 40.7 & \cellcolor{tabred!9} 58.9 &  & \cellcolor{tabred!8} 59.3 & \cellcolor{tabred!16} 55.5 & \cellcolor{tabred!12} 57.2 &  & \cellcolor{tabred!43} 42.4 & \cellcolor{tabred!55} 26.3 & \cellcolor{tabred!55} 35.6 &  & \cellcolor{tabred!31} 47.7 \\
MAI-UI-2B & 33.5 &  & \cellcolor{tabred!13} 27.5 & \cellcolor{tabred!3} 31.8 &  & \cellcolor{tabred!14} 26.7 & 34.3 &  & \cellcolor{tabred!5} 30.9 & \cellcolor{tabred!15} 26.3 & \cellcolor{tabred!13} 27.5 &  & \cellcolor{tabred!9} 29.2 & \cellcolor{tabred!16} 25.8 & \cellcolor{tabred!7} 30.1 &  & \cellcolor{tabred!9} 29.0 \\
MAI-UI-8B & 53.4 &  & \cellcolor{tabred!28} 40.3 & \cellcolor{tabred!3} 51.7 &  & \cellcolor{tabred!14} 47.0 & 54.2 &  & 52.5 & \cellcolor{tabred!7} 50.0 & \cellcolor{tabred!4} 51.3 &  & \cellcolor{tabred!7} 50.0 & \cellcolor{tabred!19} 44.5 & \cellcolor{tabred!16} 45.8 &  & \cellcolor{tabred!10} 48.7 \\
Qwen3-VL-4B-Thinking & 57.2 &  & \cellcolor{tabred!26} 45.3 & \cellcolor{tabred!15} 50.0 &  & \cellcolor{tabred!35} 41.1 & \cellcolor{tabred!15} 50.0 &  & \cellcolor{tabred!14} 50.8 & \cellcolor{tabred!7} 53.8 & \cellcolor{tabred!3} 55.5 &  & \cellcolor{tabred!20} 47.9 & \cellcolor{tabred!30} 43.2 & \cellcolor{tabred!23} 46.6 &  & \cellcolor{tabred!18} 48.4 \\
Qwen3-VL-8B-Thinking & 62.7 &  & \cellcolor{tabred!7} 59.3 & \cellcolor{tabred!3} 61.0 &  & \cellcolor{tabred!30} 48.7 & 61.9 &  & \cellcolor{tabgreen!7} 64.4 & \cellcolor{tabred!3} 61.0 & \cellcolor{tabred!5} 60.2 &  & \cellcolor{tabred!18} 54.2 & \cellcolor{tabred!8} 58.9 & \cellcolor{tabred!55} 36.9 &  & \cellcolor{tabred!13} 56.6 \\
UI-TARS-1.5-7B & 29.7 &  & \cellcolor{tabred!3} 28.0 & \cellcolor{tabred!5} 27.1 &  & \cellcolor{tabred!17} 21.6 & \cellcolor{tabred!14} 22.9 &  & \cellcolor{tabred!26} 17.8 & \cellcolor{tabred!18} 21.2 & \cellcolor{tabred!18} 21.2 &  & \cellcolor{tabred!13} 23.7 & \cellcolor{tabred!35} 13.6 & \cellcolor{tabred!19} 20.8 &  & \cellcolor{tabred!16} 21.8 \\
\bottomrule
\end{tabular}%
}
\caption{Success rates (\%) of GUI agents on original tasks and under each trap subcategory of \ours. Cell color intensity indicates performance drop from the original baseline. Abbreviations: \textbf{Ext.I.}~=~External Interruption, \textbf{Vis.O.}~=~Visual Obscuration, \textbf{Tmp.C.}~=~Temporal Conflict, \textbf{Vis.H.}~=~Visual Hallucination, \textbf{Grd.E.}~=~Grounding Error, \textbf{Typ.M.}~=~Type Mismatch, \textbf{Int.D.}~=~Intent Deviation, \textbf{St.DL.}~=~State Deadlock, \textbf{Ctx.D.}~=~Context Disruption.}
\label{tab:overall_success_rate}
\end{table*}

\paragraph{Dynamic Trap Construction.}
To simulate authentic dynamic anomalies, we build a runtime pipeline that injects traps during agent execution, guided by the \textit{STAR} taxonomy.
Specifically, as shown in Figure~\ref{fig:pipeline}, for \textbf{state traps $S_{\text{trap}}$}, we directly modify the emulator's state. This is achieved by applying image-level edits via native UI elements on screenshots or broadcasting a customized APK that triggers pop-ups, which support both predefined templates and real-time LLM-generated contents. The changed observations are recorded in the agent's history.
For \textbf{thinking traps $T_{\text{trap}}$}, we substitute single-step observations with incorrect ones to induce hallucinations or temporal conflicts, simulating cognitive challenges. Notably, the changed observation is excluded from the history, while the agent's induced output is permanently logged. Consequently, the agent must navigate subsequent states burdened by its own hallucinated context, which necessitates autonomous recovery from internal thinking errors.
For \textbf{action traps $A_{\text{trap}}$}, we intercept the agent's output to dynamically override coordinates or action types before execution, with the edited physical actions logged.
For \textbf{round traps $R_{\text{trap}}$}, we hijack the true execution flow across multiple steps, utilizing predefined trajectories or runtime execution history segments to disrupt multi-step reasoning and execution. The entirety of these prolonged disruptions is recorded in the history.
All the implementation adheres to two principles. First, we enforce \textit{dynamic intervention}, which means perturbations are injected stochastically along the trajectory rather than applied to static trajectories. Second, we guarantee \textit{solvability preservation}, indicating that the injection never renders the user instruction impossible to complete, so that any performance degradation is attributable solely to the agent's inadequate robustness to runtime anomalies.
Full implementation details are provided in the Appendix~\ref{app:implementation}.

\subsection{Data analysis}
\label{sec:data_release}
\paragraph{Statistics.}
Our benchmark pairs the \tasknum original tasks with their adversarial counterparts, which all have stochastic elements in instructions and environments. The benchmark supports the evaluation of all four-layer and ten-subcategory anomalies across every task, enabling a granular assessment of agent robustness. We additionally expose configurable parameters for trigger time and injection frequency, allowing users to adjust task difficulty as detailed in Appendix~\ref{app:implementation}.

\paragraph{Human Validation.}
To validate the reliability of \ours, we conduct a human study on all tasks across ten subcategories. Expert annotators are instructed to verify three criteria: (1) the solvability of the original task, (2) the preserved solvability after perturbation injection, and (3) the validity of the trap (\emph{i.e.,} whether it plausibly occurs in real mobile environments). Across the validated sample, $91\%$ of tasks satisfy all three criteria. Failing tasks are revised before inclusion.

\section{Main Experiments}
\subsection{Experiment Setup}
\paragraph{Metrics in Evaluation.}
Following previous work~\cite{xu2024androidlab,rawles2024androidworld}, we use task rule-based success rate as the evaluation metric with Pass@3 and unified settings as detailed in Appendix~\ref{app:implementation}. We also demonstrate the performance drop compared to the original environment without traps to reflect the robustness against perturbations.

\paragraph{Evaluated Models.}
We evaluate a wide range of multimodal models  which possess agent capabilities as follows: \textit{(1) Thinking Models} including
Gemini-3-Pro~\cite{google2025gemini3}, Claude-Sonnet-4.6~\cite{anthropic2026sonnet46}, GPT-5.4, and GPT-5.4-Mini~\cite{openai2026gpt54}.
UI-TARS-1.5-7B~\cite{qin2025ui}; MAI-UI~\cite{zhou2025maiuitechnicalreportrealworld}, for which we include 2B and 8B versions; GUI-Owl-7B~\cite{ye2025mobile}; GUI-Owl-1.5-Think~\cite{xu2026mobileagentv35multiplatformfundamentalgui}, for which we include 32B and 8B versions; Qwen3-VL-Thinking~\cite{qwen3technicalreport}, for which we include 4B and 8B versions.
\textit{(2) Instruct Models} including
GUI-Owl-1.5-Instruct~\cite{xu2026mobileagentv35multiplatformfundamentalgui}, for which we include 8B and 32B versions; Qwen3-VL-Instruct~\cite{qwen3technicalreport}, for which we include 4B and 8B versions.
Notably, there are some special considerations regarding the inference method of GPT models and thinking-layer trap design for instruct models as detailed in Appendix~\ref{app:implementation}.

\begin{figure*}[htbp]
    \centering
    \includegraphics[width=\linewidth]{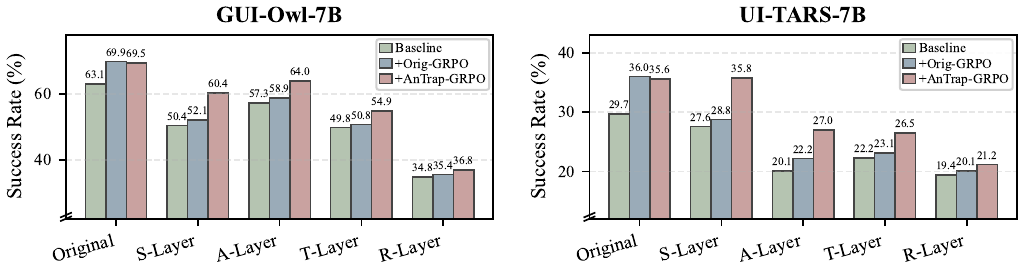}
    \caption{Layer-level performance comparison across training regimes. Grouped bars show the results of baseline, original-env GRPO, and \ours-env GRPO for each trap layer.}
    \label{fig:grpo_layer}
\end{figure*}

\paragraph{Baseline and Human-level Performance.}
We utilize original environment as baseline without traps to evaluate agent robustness based on performance degradation. We also assess human performance on \ours. Human annotators receive prior training and complete tasks relying solely on visual screen observations and environmental interactions. To prevent data leakage from identical original set, we randomly assign tasks without perturbation hints and ensure each evaluator completes exactly one task per subcategory. Note that some categories are not applicable to humans.

\subsection{Experiment Results}
\paragraph{\ours presents substantial challenges for current agentic models.}
The results reveal a universal vulnerability among current GUI agents when exposed to runtime anomalies. Across all evaluated models, ranging from proprietary models to open source ones, the average success rate under trapped conditions consistently falls below the original baseline performance. Even the most capable models, such as Claude-Sonnet-4.6 and GUI-Owl-1.5-32B-Think, experience notable performance degradations, dropping from 74.2\% to 66.5\% and 69.5\% to 62.4\% respectively. This uniform decline shows that existing agents are primarily optimized for ideal execution paths and lack the robustness required to deal with unexpected anomalies.

\paragraph{Models struggle more with contextual anomalies than single-layer ones.}
The four trap layers differ markedly in difficulty.
Based on the subcategory results in Table~\ref{tab:overall_success_rate}, most models experience a significant performance degradation under R-layer traps. Conversely, their performance is relatively better preserved under single-step anomalies, including A-layer traps, Visual Obscuration in S-layer, and Visual Hallucination in T-layer. This observation suggests that models can tolerate simple, immediate disturbances better but struggle with complex scenarios requiring cross-step reasoning and long-term dependencies, including Round-layer tasks and Temporal Conflicts in T-layer.
Additionally, the substantial performance drop caused by External Interruption indicates a weak capacity for addressing unexpected events. We hypothesize that this vulnerability stems from a lack of unexpected anomaly data during model training, as well as the agents' tendency to prioritize the current observation over long-horizon historical context, which we further explore more in Section~\ref{sec:exp_strategy}.

\paragraph{Reasoning capabilities do not equate to robustness.}

While reasoning mechanisms raise the general performance in clean environments, they do not improve robustness against runtime anomalies.
As illustrated in Table~\ref{tab:overall_success_rate}, thinking models establish higher original baselines than their standard instruct versions. For example, Qwen3-VL-8B-Thinking achieves an original success rate of 62.7\% compared to the 52.1\% of its instruct version. However, when evaluated in the \ours, the absolute magnitude of performance degradation for thinking models remains comparable to, and even occasionally exceeds, that of instruct models. Quantitatively, Qwen3-VL-8B-Thinking drops 6.1 points under traps (62.7\%$\rightarrow$56.6\%), essentially matching its Instruct counterpart's 5.8-point drop (52.1\%$\rightarrow$46.3\%). GUI-Owl-1.5-32B-Think loses 7.1 points (69.5\%$\rightarrow$62.4\%), \emph{more} than the 5.8 points lost by GUI-Owl-1.5-32B-Instruct (68.2\%$\rightarrow$62.4\%).
This indicates that current thinking paradigms alone are insufficient to bridge the fundamental gap in robustness.

\section{Exploring Strategies to Enhance GUI Agents Robustness}
\label{sec:exp_strategy}
In this section, we conduct Group Relative Policy Optimization (GRPO) experiments in both the original and the adversarial environments to investigate how RL in different environments influences GUI agent performance. We first present preliminary knowledge of GRPO (Section~\ref{sec:exp_pre}). Then we introduce the experiment details of RL in both the original environment (Section~\ref{sec:exp_rlo}) and the \ours environment (Section~\ref{sec:exp_rla}).

\subsection{Preliminary}
\label{sec:exp_pre}

GRPO is widely utilized to train agents on trajectory-level in dynamic environments. Formally, for a given task, the agents rollout a group of $G$ trajectories, which are subsequently evaluated by a reward function to determine their quality. GRPO optimizes the policy by deriving advantage values through relative comparisons among the group without a separate critic model. The simplified GRPO policy gradient objective in training is formulated as shown in Equation~\ref{eq:grpo_obj}:
\begin{equation}
\resizebox{0.89\linewidth}{!}{$
\mathcal{J}_{GRPO}(\theta) = \mathbb{E} \left[ \frac{1}{G} \sum_{i=1}^G \left( \frac{1}{|o_i|} \sum_{t=1}^{|o_i|} \rho_{i,t}(\theta) \right) A_i \right]
$}
\label{eq:grpo_obj}
\end{equation}
where $\rho_{i,t}(\theta) = \frac{\pi_\theta(o_i(t) | o_{i,<t})}{\pi_{\text{old}}(o_i(t) | o_{i,<t})}$ is the token-level probability ratio for the $t$-th token in completion $o_i$ given its preceding context $o_{i,<t}$. This ratio is practically clipped by $\epsilon$ to bound policy updates. Because the environment feedback is evaluated at the trajectory level, all tokens within the same completion $o_i$ share a uniform sequence-level relative advantage $A_i$, bypassing fine-grained token-wise return estimations. The advantage normalizes the trajectory rewards:
$A_i = \frac{r_i - \text{mean}(\mathbf{r})}{\text{std}(\mathbf{r})}$,
where $r_i$ is the individual sequence reward and $\mathbf{r}$ denotes the reward set of the sampled group.
Following this paradigm, we conduct GRPO training on UI-TARS-1.5-7B and GUI-Owl-7B in both the original and the adversarial environment. This comparative setup allows us to systematically observe how GRPO in different environments influences the general performance and the dynamic robustness against runtime anomalies of the agents.

\subsection{GRPO in Original Environment}
\label{sec:exp_rlo}
We train UI-TARS-1.5-7B and GUI-Owl-7B using sequence-level GRPO in the original environment without any injection. For each task, we sample $G=8$ rollout trajectories and assign a binary reward based on rule-based task completion. More experiment details are provided in Appendix~\ref{app:implementation}.

\paragraph{Results on Original Tasks.}
As shown in Table~\ref{tab:grpo_subcategory} and Figure~\ref{fig:grpo_layer}, GRPO in the original environment yields consistent improvements on standard task completion. GUI-Owl-7B improves from 63.1\% to 69.9\%, and UI-TARS-7B from 29.7\% to 36.0\%, confirming that online RL effectively enhances general task execution capabilities.

\paragraph{Results on Adversarial Tasks.}
The robustness gains against runtime anomalies are marginal. As shown in Figure~\ref{fig:grpo_layer}, at the layer level, A-layer exhibits some improvement (+1.6\% for GUI-Owl, +2.1\% for UI-TARS). We attribute this to the better action execution patterns learned during standard training, which offer implicit robustness to action-level perturbations. S-layer receives less benefit (+1.2\%$\sim$1.7\%), while T-layer and R-layer gain negligible improvement (under +1.0\% across most subcategories).
These results suggest that while reinforcement learning in the original environment improves the agent's baseline performance, it fails to strengthen the robustness and recovery capabilities needed to handle runtime anomalies.

\subsection{GRPO in \ours Environment}
\label{sec:exp_rla}
We conduct ten independent GRPO runs, each dedicated to a single trap subcategory. For each run, training tasks are sampled with only the corresponding subcategory's traps injected during rollout, while other settings remain consistent with the original one. Consequently, each subcategory's \textit{+Trap} entry in Table~\ref{tab:grpo_subcategory} reports the model trained and evaluated on the same subcategory.

\paragraph{Results on Original Tasks.}
Despite training each model on a single trap subcategory, all ten runs yield comparable improvements on standard tasks. Averaging across the ten subcategory-specific models, GUI-Owl-7B improves 6.4\% and UI-TARS-7B improves 5.9\%. This demonstrates that training in adversarial environments yields general GUI task execution capabilities comparable to those achieved through standard training.

\begin{table}[t]
\centering
\small
\setlength{\tabcolsep}{2.5pt}
\renewcommand{\arraystretch}{1.15}
\scalebox{0.82}{
\begin{tabular}{ll ccc ccc}
\toprule
& & \multicolumn{3}{c}{\textbf{GUI-Owl-7B}} & \multicolumn{3}{c}{\textbf{UI-TARS-7B}} \\
\cmidrule(lr){3-5} \cmidrule(lr){6-8}
\textbf{Layer} & \textbf{Subcategory} & \textbf{Base} & \textbf{+Orig} & \textbf{+Trap} & \textbf{Base} & \textbf{+Orig} & \textbf{+Trap} \\
\midrule
-- & Original Task & 63.1 & 69.9\gmark & 69.5\gmark & 29.7 & 36.0\gmark & 35.6\gmark \\
\midrule
\multirow{2}{*}{S} & Ext. Interruption & 43.2 & 45.3 & 54.2\gmark & 28.0 & 29.2 & 36.4\gmark \\
& Vis. Obscuration & 57.6 & 58.9 & 66.5\gmark & 27.1 & 28.4 & 35.2\gmark \\
\midrule
\multirow{2}{*}{T} & Tmp. Conflict & 40.7 & 41.9 & 45.8\gmark & 21.6 & 22.5 & 25.8 \\
& Vis. Hallucination & 58.9 & 59.7 & 64.0\gmark & 22.9 & 23.7 & 27.1 \\
\midrule
\multirow{3}{*}{A} & Grounding Err. & 59.3 & 62.7 & 67.4\gmark & 17.8 & 20.8 & 25.4\gmark \\
& Type Mismatch & 55.5 & 58.5 & 64.0\gmark & 21.2 & 24.2 & 28.8\gmark \\
& Int. Deviation & 57.2 & 55.5 & 60.6 & 21.2 & 21.6 & 26.7\gmark \\
\midrule
\multirow{3}{*}{R} & State Deadlock & 42.4 & 42.8 & 44.9 & 23.7 & 24.6 & 26.3 \\
& Ctx. Disruption & 26.3 & 27.5 & 29.2 & 13.6 & 14.4 & 15.7 \\
& Loop & 35.6 & 36.0 & 36.4 & 20.8 & 21.2 & 21.6 \\
\bottomrule
\end{tabular}
}
\caption{Success rates (\%) under different subcategories of GRPO training regimes. \textbf{+Orig}: trained in original environment; \textbf{+Trap}: trained in \ours environment. \gmark~denotes $\geq$5\% absolute improvement over baseline.}
\label{tab:grpo_subcategory}
\end{table}

\paragraph{Results on Adversarial Tasks.}
In contrast to the standard environment training, \ours-GRPO yields differentiated robustness gains across trap categories. The models demonstrate substantial improvements against state-level perturbations (S-layer, +8.1\%$\sim$11.0\%) and most action-level execution errors (A-layer, up to +8.5\%, though Intent Deviation gains more modestly at +3.4\%$\sim$5.5\%).
In these cases, successful rollouts against anomalies provide clear learning signals for the agents to identify and adapt to specific perturbations, leading to notable robustness improvements.
However, gains diminish to a moderate level for reasoning traps (T-layer, +4.2\%$\sim$5.1\%) and become completely negligible for multi-step contextual anomalies (R-layer, under +3.0\%, with Loop traps strictly below +1.0\%).
This indicates that while single-step anomalies are learnable through adversarial exposure, complex contextual traps are difficult to overcome via RL. We attribute this to the agents' reliance and proficiency in leveraging current-step information rather than integrating broader context for next-step planning. More meta-cognitive, long-horizon self-monitoring capabilities are required for agents to escape circular or deadlocked states.

\vspace{0.55em}
\section{Conclusion}
We present \ours, a benchmark designed for systematically evaluating GUI agents' robustness against runtime anomalies in Android environments, featuring a taxonomy \textit{STAR} and a meticulously designed construction pipeline. We evaluate \modelnum models on \ours, revealing their vulnerability to dynamic traps across all layers and subcategories. We further conduct GRPO training in both original and adversarial environments to delineate the boundary between environment-learnable and reasoning-bottlenecked challenges and find that adversarial RL can improve most single-step robustness in the state and action layers, while deeper contextual traps remain hard to address.

\clearpage
\section*{Limitations}
While our work introduces a fine-grained benchmark for GUI agents robustness against runtime anomalies, we acknowledge several limitations that open avenues for future research.
First, due to the time-consuming nature of annotating and validating the online original task initialization and evaluation, we have only tested and studied \ours on \tasknum base tasks. These tasks, while diverse, are still limited and may not fully represent the wide range of common Android GUI operation scenarios globally. Future work could apply our method to a broader array of Android scenarios to enhance the generalizability of our findings.
Second, due to budget constraints and the difficulty of obtaining annotated trajectory data, we do not explore adversarial supervised fine-tuning (SFT) in our study, which has the potential to address issues related to contextual understanding~\cite{wu2025gui, chen2025perturbollava}. Future research could investigate the effectiveness of adversarial SFT in improving model performance under runtime anomalies.
Finally, our work intentionally focuses on evaluation. We designed \ours primarily as a benchmark for diagnosing model weaknesses rather than a general training solution. Developing scalable training methods to address these weaknesses remains an open research direction.

\bibliography{uiagent,rl,bench,llm,custom}

\iffullbuild
  \appendix
  \newpage
\clearpage
\section{Implementation Details}
\label{app:implementation}

\begin{table*}[!ht]
\centering
\small
\renewcommand{\arraystretch}{1.35}
\begin{tabularx}{\textwidth}{@{}l X @{}}
\toprule
\textbf{Subcategory} & \textbf{Implementation} \\ \midrule

\multicolumn{2}{c}{\textbf{Category 1: State Layer}} \\
(1) External Interruption
& Pop-ups are triggered through a dedicated \textit{TrapOverlay} APK installed on the emulator, which registers a \textit{SYSTEM\_ALERT\_WINDOW} overlay and reacts to Android Debug Bridge (ADB) broadcasts. Four pop-up types are supported: runtime permission dialogs, modal center dialogs, top notification banners, and fullscreen advertisements. Content is drawn from a curated bank of $92$ realistic Android prompts ($21$ permission, $32$ center-dialog, $31$ banner, and $8$ fullscreen-ad templates). Alternatively, the agent's own Vision-Language Model (VLM) generates content online, conditioned on the foreground-app package name; this mode falls back to the template bank on JSON-parse failure or when the foreground process is a system launcher. \\

(2) Visual Obscuration
& Pixel-level masks are applied to the captured frame through Pillow. The target region is one of three: a random clickable or editable UI element drawn from the accessibility tree, a uniformly sampled element regardless of affordance, or a random rectangle covering $10$--$30\%$ of the screen. The selected region is replaced by either a Gaussian blur with configurable kernel radius (default $r{=}15$) or a solid color patch when a color is specified. \\ \midrule

\multicolumn{2}{c}{\textbf{Category 2: Thinking Layer}} \\
(3) Temporal Conflict
& A rolling buffer of past frames is maintained throughout the episode. At trigger time the current observation passed to the model is replaced with the frame captured $N$ steps earlier (default $N{=}3$), while the unperturbed frame is still saved into the agent's persistent history. As an alternative, a random screenshot can be sampled from a pre-collected pool of unrelated app states and resized to the current resolution. \\

(4) Visual Hallucination
& For each text-bearing UI element identified via its accessibility-tree bounding box, the background color is estimated by median sampling along the bbox edges, the original region is overwritten with that color, and a misleading replacement is rendered in DejaVu Sans with the contrast-inverted color. Replacements are drawn from a curated $24$-pair antonym dictionary covering common UI verbs such as Save/Delete, Allow/Deny, and Confirm/Cancel; unrecognized labels fall back to a character-level permutation. \\ \midrule

\multicolumn{2}{c}{\textbf{Category 3: Action Layer}} \\
(5) Grounding Error
& The parsed action is intercepted between model output and ADB execution. An independent random offset $\delta_x, \delta_y$ is added to its coordinates, with magnitude uniformly sampled from $[20, 50]$ pixels and uniformly random sign. The same perturbation is applied to all touch-style actions (\textit{click}, \textit{long\_press}, \textit{double\_tap}) as well as the start of \textit{swipe}. The model's recorded intent is left unchanged so that the offset is only observable through the resulting screenshot. \\

(6) Type Mismatch
& At the same interception point, the action's type is rewritten. Touch actions are randomly remapped among the three alternatives \textit{click}, \textit{long\_press}, and \textit{double\_tap}; \textit{swipe} is reversed by swapping its start and end coordinates. The model's response text in the conversation history retains the originally intended action type, so any resulting mismatch must be detected from the post-execution screenshot. \\

(7) Intent Deviation
& Before execution, the predicted action is intercepted and changed to a different operation. For touch actions, we change the click coordinates to target a different UI element on the screen, preferring nearby buttons with opposite meanings (e.g., clicking "Cancel" instead of "Confirm"). For \textit{type\_text} actions, the input text is replaced with wrong text from a dictionary or random characters. Importantly, the agent's history still records its original reasoning and intended action. This creates a mismatch between what the agent planned and what actually happened, forcing the agent to notice the mistake from the next screenshot. \\ \midrule

\multicolumn{2}{c}{\textbf{Category 4: Round Layer}} \\
(8) State Deadlock
& At the execution hook, a block signal is returned for $k$ consecutive steps, where $k$ is uniformly sampled from $[1, \lfloor T/4 \rfloor]$ and $T$ is the original step budget. During this window every ADB command is silently dropped, so the screenshot returned to the agent is identical to the previous one regardless of the action chosen. \\

(9) Context Disruption
& Immediately before the agent's step begins, a single ADB keyevent is injected. Two subtypes are supported: \textit{KEYCODE\_HOME}, which returns the device to the launcher, and \textit{KEYCODE\_APP\_SWITCH}, which opens the recent-apps view. The hook then waits one second for the UI to settle so the next screenshot reflects the disrupted context. \\

(10) Loop
& At a fixed cadence (every $N$ steps, $N{=}2$ by default), \textit{KEYCODE\_BACK} is injected before the agent's step, forcing repeated backward navigation regardless of the current task progress. The loop terminates once the per-episode trap budget has been spent. \\
\bottomrule
\end{tabularx}
\caption{Detailed implementation of the ten dynamic trap operators in the \ours benchmark. Each row specifies the concrete mechanism, content sources, and parameters used to realize the corresponding trap during evaluation.}
\label{tab:trap_implementation_details}
\end{table*}

\subsection{\ours Construction}
\label{sec:ours-construction}
\paragraph{Original Task Augmentation.}
We expand the original AndroidWorld suite of $116$ tasks to \tasknum tasks by manually adding $120$ new tasks spanning the existing application families. Every new task is manually verified to confirm that it initializes a valid starting state, can be completed through the standard Android UI, and is correctly judged by its evaluator, so the augmented suite preserves the solvability and technical soundness of the original benchmark. We release all task definitions and evaluators in our anonymous codebase.

\paragraph{Framework.}
\ours augments a standard Android environment with an adversarial layer that injects the anomalies defined in our \textit{STAR} taxonomy into the agent's interaction loop, while leaving the underlying tasks, evaluators, and the agent's reasoning logic unchanged. The framework consists of three conceptual components: a configuration that specifies which anomaly is instantiated and at which layer, a central controller that orchestrates injection during execution, and four layer-specific modules that realize the ten trap operators of the taxonomy. Because intervention occurs only at the interface between the agent and the environment, instrumenting an existing agent does not require modifying its inference logic.

\paragraph{Intervention Points.}
\ours injects perturbations at four points along the agent's interaction loop, each aligned to one layer of the taxonomy. \textit{State layer} perturbations modify the observation as it is captured, persistently distorting the input on which all downstream reasoning depends. \textit{Thinking layer} perturbations instead act on the observation only at decision time, transiently presenting an old or label-tampered view to the model. \textit{Action layer} perturbations are injected between the agent's output action and its dispatch to the device, offsetting coordinates, remapping the action type, or in the case of State Deadlock, suppressing dispatch entirely to simulate an unresponsive device. \textit{Round-layer} perturbations operate at the trajectory level, supporting traps such as Context Disruption and Loop. When no trap matches at the current step, every intervention reduces to a transparent pass-through, so injection in \ours imposes negligible overhead on clean runs except for the time cost for recovery steps.

\paragraph{Trigger Mechanism.}
In our experiment, at each step, a trap is injected independently with probability $p_{\text{trap}}=0.16$, subject to a per-episode cap of $n_{\max}=1$, so that every reported episode contains exactly one trap event. To preserve enough subsequent steps for the agent to recover, traps are restricted to the first $80\%$ of the original step budget, and the total budget is extended by a factor of $1.15\times$. Episodes in which no trap is injected within ten retries are excluded from reporting. In practice, an average of $1.3$ attempts suffices to inject a trap. All stochastic decisions including trap-type sampling, inject choice, blur regions, and coordinate offsets are governed by a fixed random seed of $42$.

\paragraph{History Policy.}
The ten trap operators fall into two modes, distinguished by whether the perturbation enters the agent's persistent interaction history. \emph{Direct modifications} perturb the environment as the agent sees it. The altered state is recorded in the log as the state of the current step and remains visible in subsequent reasoning, so we measure whether the agent can recognize and recover from a real change in the world.
\emph{Induced modifications}, in contrast, only record the agent's misled outputs including offset and remapped actions, without the deceptive observations or the action-hijacking events.
This means the agent receives a transiently corrupted observation at decision time, or its output action is silently dropped before execution. Because the agent's perceptions and stated intent are logged exactly as in a clean run, any resulting failure is self-induced, caused by the agent's own decisions under momentarily misleading inputs. Thus, the trap measures self-correction rather than tolerance to external perturbation.

\subsection{Benchmark Reproduction}
\paragraph{Hardware and Models.}
Inference runs on a server with Intel Xeon CPUs and four NVIDIA~A40 GPUs (48~GB VRAM each).
Open-weight VLMs are served by vLLM with tensor-parallel size $4$, exposed via an OpenAI-compatible API. We use a context length of $32{,}768$ tokens and at most $8$ images per prompt.
Other settings follow their official model-card defaults in huggingface or technical reports.
For API-based models (Claude, GPT and Gemini), we use temperature $1.0$, a $4{,}096$-token output cap, and a $8$-image history buffer.

\paragraph{Evaluation Settings.}
We run $8$ Android emulators in parallel in a Docker container emulating a Pixel~6 on Android API Level~33, with paired console and gRPC ports for network communication. For reproducibility, we fix the task-initialization seed at $30$ and the trap seed at $42$. Each evaluated episode contains exactly one trap event, with the trigger schedule determined by $p_{\text{trap}}=0.16$ and the retry policy in the previous paragraph. The per-step prompts used for each model are listed in Figures~\ref{fig:prompt_popup} to Figures~\ref{fig:prompt_guiowl_qwen3vl_gemini3}. For the human-level evaluation, we recruit five annotators and two authors to independently complete the tasks, and pay \$10 per annotator.

\paragraph{Special Considerations.}
Notably, for the GPT models, following the approach of AndroidLab~\cite{xu2024androidlab}, we adopt a Set of Marks (SoM) inference style without a grounding requirement. We therefore exclude grounding error traps for GPT models.
For instruct models, although they do not follow the ReAct reasoning framework, their official prompts are designed to guide them to first output a short description of the action before generating the tool call. Therefore, the perturbations injected into the T-layer effectively disrupt the action description phase for these models.

\begin{table}[t]
\centering
\small
\setlength{\tabcolsep}{5pt}
\begin{tabular}{lll}
\toprule
Component & Hyperparameter & Value \\
\midrule
Data & Max Prompt Len. & 32768 \\
Data & Max Response Len. & 512 \\
Data & Train Batch Size & 8 \\
Actor/Policy & Parallelism & FSDP2 \\
Actor/Policy & Micro Batch/GPU & 1 \\
Actor/Policy & LR & 1e-6 \\
Actor/Policy & Grad. Clip & 1.0 \\
Actor/Policy & Clip Ratio & 0.2 \\
Sampling & Temp. & 1.0 \\
Sampling & Max Tokens & 512 \\
Sampling & Max Turns & 50 \\
Sampling & Max Pixels & 1270180 \\
Sampling & Min Pixels & 256 \\
\bottomrule
\end{tabular}
\caption{Main hyperparameters in GRPO.}
\label{tab:hyperparam}
\end{table}

\subsection{GRPO Reproduction}
We conduct full-parameter GRPO on 2 nodes of $8$ NVIDIA A100 GPUs (80GB), leveraging a customized \textsc{verl} framework~\cite{sheng2024hybridflow}. We construct our training task pool by leveraging original tasks with randomized parameters. Following a manual verification process to ensure feasibility and strictly exclude overlaps with the test set, we compile a final dataset of 600 training tasks. To optimize throughput, we scale parallel emulators to match the training batch size, accelerate rollouts via vLLM~\cite{kwon2023efficient}, and employ FSDP2 with BF16 mixed precision. The policy has access to the interaction history throughout training. Unless otherwise specified, these settings apply to all GRPO experiments, with comprehensive hyperparameters detailed in Table~\ref{tab:hyperparam}.

\section{Pilot Study for Taxonomy Development}
\label{app:pilot_work}

\paragraph{Task Collection and Execution.}
To establish a grounded taxonomy of real-world agent failures, we conduct a large-scale pilot study prior to benchmark construction. We predefined 600 daily mobile tasks spanning common usage scenarios (e.g., messaging, navigation, media playback, settings configuration, and shopping) across 30+ popular Android applications. We deployed GUI agents including UITARS-1.5-7B, GUI-Owl-7B, and MAIUI-8B on a cloud server and connected to physical Android devices via ADB, executing each task under authentic network conditions and system states. All trajectories, including screenshots, model reasoning, and executed actions at each step, are recorded for subsequent analysis.

\paragraph{Annotation Protocol and Taxonomy Derivation.}

Two of our authors serve as annotators in this stage, independently reviewing the full set of recorded trajectories using a custom web-based annotation interface (Figure~\ref{fig:prior_html}). Annotators examined every trajectory step-by-step, not only those from failed tasks. This allows us to capture cases where the agent eventually completed the task but encountered intermediate errors requiring recovery.
We randomly sampled trajectories from 150 tasks for open coding, where annotators marked the presence of issues at each step and assigned an initial error category with notes. Through iterative discussion and annotation analysis on the remaining 450 tasks, we progressively merged and refined the categories, ultimately forming a four-layer ten-subcategory taxonomy. This ensures that each category reflects recurring, empirically grounded failure modes rather than hypothetical edge cases.

\section{Qualitative Results of \ours}
We present qualitative examples of GUI agents' trajectories under different dynamic traps and different settings in the \ours benchmark as shown in Figure~\ref{fig:example0} to~\ref{fig:example4}.
\label{sec:qual_examples}

\section{Mixed-Trap GRPO}
\label{app:random_trap}
\begin{table}[h]
\centering
\small
\setlength{\tabcolsep}{2.5pt}
\renewcommand{\arraystretch}{1.15}
\scalebox{0.82}{
\begin{tabular}{ll cc cc}
\toprule
& & \multicolumn{2}{c}{\textbf{GUI-Owl-7B}} & \multicolumn{2}{c}{\textbf{UI-TARS-7B}} \\
\cmidrule(lr){3-4} \cmidrule(lr){5-6}
\textbf{Layer} & \textbf{Subcategory} & \textbf{Base} & \textbf{+Mixed} & \textbf{Base} & \textbf{+Mixed} \\
\midrule
-- & Original Task & 63.1 & 69.1 & 29.7 & 35.6 \\
\midrule
\multirow{2}{*}{S} & Ext. Interruption & 43.2 & 46.2 & 28.0 & 30.1 \\
& Vis. Obscuration & 57.6 & 59.7 & 27.1 & 29.2 \\
\midrule
\multirow{2}{*}{T} & Tmp. Conflict & 40.7 & 42.4 & 21.6 & 22.9 \\
& Vis. Hallucination & 58.9 & 60.2 & 22.9 & 24.2 \\
\midrule
\multirow{3}{*}{A} & Grounding Err. & 59.3 & 62.3 & 17.8 & 20.3 \\
& Type Mismatch & 55.5 & 58.1 & 21.2 & 23.7 \\
& Int. Deviation & 57.2 & 55.5 & 21.2 & 21.2 \\
\midrule
\multirow{3}{*}{R} & State Deadlock & 42.4 & 43.2 & 23.7 & 25.0 \\
& Ctx. Disruption & 26.3 & 27.1 & 13.6 & 14.4 \\
& Loop & 35.6 & 36.0 & 20.8 & 21.2 \\
\bottomrule
\end{tabular}
}
\caption{Results of mixed-trap GRPO training, where all ten trap subcategories are randomly injected.}
\label{tab:grpo_random}
\end{table}

In the main experiments (Section~\ref{sec:exp_rla}), we adopt a per-subcategory training protocol that yields ten specialist models per base agent, each trained on a single trap subcategory. Here we report a supplementary experiment using \textbf{mixed-trap training}, where one model is trained per base agent and all ten trap subcategories are randomly sampled during each rollout at an equal probability. The training configuration is otherwise identical with 600 tasks, $G=8$ rollouts, binary reward.
As shown in Table~\ref{tab:grpo_random}, mixed-trap training yields improvements on original tasks comparable to per-subcategory training. However, the robustness gains across trap subcategories are substantially diluted, with improvements roughly on par with original-environment GRPO (Section~\ref{sec:exp_rlo}). This indicates that concentrated, subcategory-specific adversarial exposure is critical for meaningful robustness gains, as random mixing distributes the learning signal too sparsely across diverse recovery strategies, and may need more training samples to converge to a robust policy.

\onecolumn
\clearpage

\begin{figure*}[!htbp]
    \centering
    \begin{tcolorbox}[
        title=Prompt for External Interruption Pop-up Content Generation,
        colback=gray!2!white,
        colframe=gray!40!black,
        fonttitle=\bfseries,
        arc=5pt,
        boxrule=1pt,
        colbacktitle=gray!15!white,
        coltitle=black,
    ]
\begin{lstlisting}[
    basicstyle=\ttfamily\small,
    breaklines=true,
    breakatwhitespace=false,
    breakindent=0pt,
    columns=fullflexible,
    keepspaces=true,
    showstringspaces=false,
    upquote=true,
    extendedchars=true,
    inputencoding=utf8,
    aboveskip=0pt,
    belowskip=0pt,
    literate=%
        {—}{{\textemdash}}1%
        {–}{{\textendash}}1%
        {‘}{{`}}1 {’}{{'}}1%
        {“}{{``}}1 {”}{{''}}1%
        {•}{{\textbullet}}1%
        {→}{{$\rightarrow$}}1%
        {…}{{\ldots}}1,
]
# === pop-up type 1: Permission dialog ===
Generate a realistic Android permission request dialog for the app '{current_app}'.

Requirements:
- title: 3-6 words, specific to the app (e.g., "Camera Access Required")
- message: 10-20 words, explain why the permission is needed
- button1: positive action (e.g., "Allow", "Enable")
- button2: negative action (e.g., "Deny", "Don't Allow")

Output JSON only:
{"title": "...", "message": "...", "button1": "...", "button2": "..."}


# === pop-up type 2: Center modal dialog ===
Generate a realistic Android dialog pop-up that might appear while using '{current_app}'.

Requirements:
- title: 3-5 words, dialog purpose (e.g., "Update Available", "Sign In Required")
- message: 15-25 words, detailed information about the dialog

Output JSON only:
{"title": "...", "message": "..."}


# === pop-up type 3: Top notification banner ===
Generate a realistic Android notification banner that might appear while using '{current_app}'.

Requirements:
- title: 2-4 words, notification type (e.g., "New Message", "Update Available")
- message: 8-15 words, specific notification content

Output JSON only:
{"title": "...", "message": "..."}


# === pop-up type 4: Fullscreen advertisement ===
Generate a realistic full-screen advertisement text for an app that might appear while using '{current_app}'.

Requirements:
- title: 5-10 words, catchy promotional text (e.g., "LIMITED TIME OFFER - 50% OFF TODAY ONLY!")

Output JSON only:
{"title": "..."}
\end{lstlisting}
    \end{tcolorbox}
    \caption{Prompts used by the agent's own VLM to generate External Interruption pop-up content, contextualized on the foreground-app package name \textit{\{current\_app\}}. The four templates correspond to the four pop-up types rendered by the \textit{TrapOverlay} APK; on JSON-parse failure or when \textit{\{current\_app\}} resolves to a system launcher, the framework falls back to the curated $92$-entry template bank described in Section~\ref{sec:taxonomy}.}
    \label{fig:prompt_popup}
\end{figure*}

\begin{figure*}[!htbp]
    \centering
    \begin{tcolorbox}[
        title=System Prompt for Claude-Sonnet,
        colback=brown!3!white,
        colframe=brown!50!black,
        fonttitle=\bfseries,
        arc=5pt,
        boxrule=1pt,
        colbacktitle=brown!25!white,
        coltitle=black,
    ]
\begin{lstlisting}[
    basicstyle=\ttfamily\small,
    breaklines=true,
    breakatwhitespace=false,
    breakindent=0pt,
    columns=fullflexible,
    keepspaces=true,
    showstringspaces=false,
    upquote=true,
    extendedchars=true,
    inputencoding=utf8,
    aboveskip=0pt,
    belowskip=0pt,
    literate=%
        {—}{{\textemdash}}1%
        {–}{{\textendash}}1%
        {‘}{{`}}1 {’}{{'}}1%
        {“}{{``}}1 {”}{{''}}1%
        {•}{{\textbullet}}1%
        {→}{{$\rightarrow$}}1%
        {…}{{\ldots}}1,
]
You are an AI agent that operates an Android phone by looking at screenshots and choosing one action per turn. Your job is to make steady progress toward the user's goal: never repeat a stuck action, never invent UI elements that aren't visible, and finish with action=terminate (status="success" or "failure") once the goal is achieved or impossible.

# Coordinate system
All coordinates are normalized to a 0-999 by 0-999 grid that maps to the entire visible screenshot. (0, 0) is the top-left corner; (999, 999) is the bottom-right. Aim the cursor at the center of the target element, not its edge.

# Action vocabulary
You may emit exactly one of the following actions per turn:
- click — tap a single point.
  arguments: {"action": "click", "coordinate": [x, y]}
- long_press — press and hold a point for `time` seconds.
  arguments: {"action": "long_press", "coordinate": [x, y], "time": <seconds>}
- swipe — drag from one point to another (use for scrolling lists, dismissing sheets, opening the app drawer, etc.).
  arguments: {"action": "swipe", "coordinate": [x1, y1], "coordinate2": [x2, y2]}
- type — type text into the currently focused input box. The text appears as-is; the IME does not interpret tabs or newlines.
  arguments: {"action": "type", "text": "<string>"}
- key — issue a hardware/keyevent (adb keyevent name, e.g. "volume_up", "power", "clear").
  arguments: {"action": "key", "text": "<keyevent>"}
- system_button — press one of the soft system buttons.
  arguments: {"action": "system_button", "button": "Back" | "Home" | "Menu" | "Enter"}
- open — launch an app by name. The name MUST be from the canonical app list at the bottom of this prompt.
  arguments: {"action": "open", "text": "<app name>"}
- wait — pause and let the UI settle (use after a navigation that takes a moment to load).
  arguments: {"action": "wait", "time": <seconds>}
- answer — submit a textual answer to the user's question. Use this for tasks whose goal is information retrieval.
  arguments: {"action": "answer", "text": "<answer>"}
- terminate — end the task. Always emit this once the goal is fulfilled or judged infeasible.
  arguments: {"action": "terminate", "status": "success" | "failure"}

# How to think
Before acting, briefly reason about: (1) what is on the screen now, (2) whether your previous action achieved its intended effect, and (3) what single concrete step moves the goal forward. Keep reasoning under ~80 words; do not narrate every UI element.
If the prior action did not have the expected effect (e.g. a pop-up is still there, the keyboard didn't appear, the app didn't launch), do NOT immediately retry the same action. Diagnose first: dismiss the pop-up, scroll, switch apps, or pick a different target.

# Response format
Respond every step with EXACTLY two XML blocks, in this order, and nothing else:
<reasoning>
[Brief reasoning, see "How to think" above.]
</reasoning>
<tool_call>
{"name": "mobile_use", "arguments": {"action": "<action>", ...}}
</tool_call>
The <tool_call> block must contain ONLY a single valid JSON object — no markdown fences, no commentary.

# Available apps for action=open
Use these exact names (lowercase):
\end{lstlisting}
    \end{tcolorbox}
    \caption{System prompt used for Claude-Sonnet in \ours.}
    \label{fig:prompt_claudesonnet}
\end{figure*}

\begin{figure*}[!htbp]
    \centering
    \begin{tcolorbox}[
        title=System Prompt for GPT-5,
        colback=brown!3!white,
        colframe=brown!50!black,
        fonttitle=\bfseries,
        arc=5pt,
        boxrule=1pt,
        colbacktitle=brown!25!white,
        coltitle=black,
    ]
\begin{lstlisting}[
    basicstyle=\ttfamily\scriptsize,
    breaklines=true,
    breakatwhitespace=false,
    breakindent=0pt,
    columns=fullflexible,
    keepspaces=true,
    showstringspaces=false,
    upquote=true,
    extendedchars=true,
    inputencoding=utf8,
    aboveskip=0pt,
    belowskip=0pt,
    literate=%
        {—}{{\textemdash}}1%
        {–}{{\textendash}}1%
        {‘}{{`}}1 {’}{{'}}1%
        {“}{{``}}1 {”}{{''}}1%
        {•}{{\textbullet}}1%
        {→}{{$\rightarrow$}}1%
        {…}{{\ldots}}1,
]
You are an agent who can operate an Android phone on behalf of a user. Based on user's goal/request, you may
- Answer back if the request/goal is a question (or a chat message), like user asks "What is my schedule for today?".
- Complete some tasks described in the requests/goals by performing actions (step by step) on the phone.

When given a user request, you will try to complete it step by step. At each step, you will be given the current screenshot (including the original screenshot and the same screenshot with bounding boxes and numeric indexes added to some UI elements) and a history of what you have done (in text). Based on these pieces of information and the goal, you must choose to perform one of the action in the following list (action description followed by the JSON format) by outputing the action in the correct JSON format.
- If you think the task has been completed, finish the task by using the status action with complete as goal_status: `{{"action_type": "status", "goal_status": "complete"}}`
- If you think the task is not feasible (including cases like you don't have enough information or can not perform some necessary actions), finish by using the `status` action with infeasible as goal_status: `{{"action_type": "status", "goal_status": "infeasible"}}`
- Answer user's question: `{{"action_type": "answer", "text": "<answer_text>"}}`
- Click/tap on an element on the screen. We have added marks (bounding boxes with numeric indexes on their TOP LEFT corner) to most of the UI elements in the screenshot, use the numeric index to indicate which element you want to click: `{{"action_type": "click", "index": <target_index>}}`.
- Long press on an element on the screen, similar with the click action above, use the numeric label on the bounding box to indicate which element you want to long press: `{{"action_type": "long_press", "index": <target_index>}}`.
- Type text into a text field (this action contains clicking the text field, typing in the text and pressing the enter, so no need to click on the target field to start), use the numeric label on the bounding box to indicate the target text field: `{{"action_type": "input_text", "text": <text_input>, "index": <target_index>}}`
- Press the Enter key: `{{"action_type": "keyboard_enter"}}`
- Navigate to the home screen: `{{"action_type": "navigate_home"}}`
- Navigate back: `{{"action_type": "navigate_back"}}`
- Scroll the screen or a scrollable UI element in one of the four directions, use the same numeric index as above if you want to scroll a specific UI element, leave it empty when scroll the whole screen: `{{"action_type": "scroll", "direction": <up, down, left, right>, "index": <optional_target_index>}}`
- Open an app (nothing will happen if the app is not installed): `{{"action_type": "open_app", "app_name": <name>}}`
- Wait for the screen to update: `{{"action_type": "wait"}}`

Here are some useful guidelines you need to follow:
General:
- Usually there will be multiple ways to complete a task, pick the easiest one. Also when something does not work as expected (due to various reasons), sometimes a simple retry can solve the problem, but if it doesn't (you can see that from the history), SWITCH to other solutions.
- Sometimes you may need to navigate the phone to gather information needed to complete the task, for example if user asks "what is my schedule tomorrow", then you may want to open the calendar app (using the `open_app` action), look up information there, answer user's question (using the `answer` action) and finish (using the `status` action with complete as goal_status).
- For requests that are questions (or chat messages), remember to use the `answer` action to reply to user explicitly before finish! Merely displaying the answer on the screen is NOT sufficient (unless the goal is something like "show me ...").
- If the desired state is already achieved (e.g., enabling Wi-Fi when it's already on), you can just complete the task.
Action Related:
- Use the `open_app` action whenever you want to open an app (nothing will happen if the app is not installed), do not use the app drawer to open an app unless all other ways have failed.
- Use the `input_text` action whenever you want to type something (including password) instead of clicking characters on the keyboard one by one. Sometimes there is some default text in the text field you want to type in, remember to delete them before typing.
- For `click`, `long_press` and `input_text`, the index parameter you pick must be VISIBLE in the screenshot and also in the UI element list given to you (some elements in the list may NOT be visible on the screen so you can not interact with them).
- Consider exploring the screen by using the `scroll` action with different directions to reveal additional content.
- The direction parameter for the `scroll` action can be confusing sometimes as it's opposite to swipe, for example, to view content at the bottom, the `scroll` direction should be set to "down". It has been observed that you have difficulties in choosing the correct direction, so if one does not work, try the opposite as well.
Text Related Operations:
- Normally to select certain text on the screen: <i> Enter text selection mode by long pressing the area where the text is, then some of the words near the long press point will be selected (highlighted with two pointers indicating the range) and usually a text selection bar will also appear with options like `copy`, `paste`, `select all`, etc. <ii> Select the exact text you need. Usually the text selected from the previous step is NOT the one you want, you need to adjust the range by dragging the two pointers. If you want to select all text in the text field, simply click the `select all` button in the bar.
- At this point, you don't have the ability to drag something around the screen, so in general you can not select arbitrary text.
- To delete some text: the most traditional way is to place the cursor at the right place and use the backspace button in the keyboard to delete the characters one by one (can long press the backspace to accelerate if there are many to delete). Another approach is to first select the text you want to delete, then click the backspace button in the keyboard.
- To copy some text: first select the exact text you want to copy, which usually also brings up the text selection bar, then click the `copy` button in bar.
- To paste text into a text box, first long press the text box, then usually the text selection bar will appear with a `paste` button in it.
- When typing into a text field, sometimes an auto-complete dropdown list will appear. This usually indicating this is a enum field and you should try to select the best match by clicking the corresponding one in the list.

\end{lstlisting}
    \end{tcolorbox}
    \caption{System prompt used for GPT-5 in \ours. Since we use the SOM approach for GPT, we refer to the target element by its index in the screenshot, and the screenshot provided to GPT will have bounding boxes and index annotations for the elements. GPT needs to choose the element to operate on based on this index.}
    \label{fig:prompt_gpt5}
\end{figure*}

\begin{figure*}[!htbp]
    \centering
    \begin{tcolorbox}[
        title=System Prompt for MAI-UI,
        colback=brown!3!white,
        colframe=brown!50!black,
        fonttitle=\bfseries,
        arc=5pt,
        boxrule=1pt,
        colbacktitle=brown!25!white,
        coltitle=black,
    ]
\begin{lstlisting}[
    basicstyle=\ttfamily\small,
    breaklines=true,
    breakatwhitespace=false,
    breakindent=0pt,
    columns=fullflexible,
    keepspaces=true,
    showstringspaces=false,
    upquote=true,
    extendedchars=true,
    inputencoding=utf8,
    aboveskip=0pt,
    belowskip=0pt,
    literate=%
        {—}{{\textemdash}}1%
        {–}{{\textendash}}1%
        {‘}{{`}}1 {’}{{'}}1%
        {“}{{``}}1 {”}{{''}}1%
        {•}{{\textbullet}}1%
        {→}{{$\rightarrow$}}1%
        {…}{{\ldots}}1,
]
You are a GUI agent. You are given a task and your action history, with screenshots. You need to perform the next action to complete the task.

## Output Format
For each function call, return the thinking process in <thinking> </thinking> tags, and a json object with function name and arguments within <tool_call></tool_call> XML tags:
<thinking>
...
</thinking>
<tool_call>
{"name": "mobile_use", "arguments": <args-json-object>}
</tool_call>

## Action Space
{"action": "click", "coordinate": [x, y]}
{"action": "long_press", "coordinate": [x, y]}
{"action": "type", "text": ""}
{"action": "swipe", "direction": "up or down or left or right", "coordinate": [x, y]}
{"action": "open", "text": "app_name"}
{"action": "drag", "start_coordinate": [x1, y1], "end_coordinate": [x2, y2]}
{"action": "system_button", "button": "button_name"} # Options: back, home, menu, enter
{"action": "wait"}
{"action": "terminate", "status": "success or fail"}
{"action": "answer", "text": "xxx"}

## Note
- Write a small plan and finally summarize your next action (with its target element) in one sentence in <thinking></thinking> part.
- Available Apps: `["Camera","Chrome","Clock","Contacts","Dialer","Files","Settings","Markor","Tasks","Simple Draw Pro","Simple Gallery Pro","Simple SMS Messenger","Audio Recorder","Pro Expense","Broccoli APP","OSMand","VLC","Joplin","Retro Music","OpenTracks","Simple Calendar Pro"]`.
You should use the `open` action to open the app as possible as you can, because it is the fast way to open the app.
- You must follow the Action Space strictly, and return the correct json object within <thinking> </thinking> and <tool_call></tool_call> XML tags.
\end{lstlisting}
    \end{tcolorbox}
    \caption{System prompt used for MAI-UI in \ours.}
    \label{fig:prompt_maiui}
\end{figure*}

\begin{figure*}[!htbp]
    \centering
    \begin{tcolorbox}[
        title=System Prompt for UI-TARS,
        colback=brown!3!white,
        colframe=brown!50!black,
        fonttitle=\bfseries,
        arc=5pt,
        boxrule=1pt,
        colbacktitle=brown!25!white,
        coltitle=black,
    ]
\begin{lstlisting}[
    basicstyle=\ttfamily\small,
    breaklines=true,
    breakatwhitespace=false,
    breakindent=0pt,
    columns=fullflexible,
    keepspaces=true,
    showstringspaces=false,
    upquote=true,
    extendedchars=true,
    inputencoding=utf8,
    aboveskip=0pt,
    belowskip=0pt,
    literate=%
        {—}{{\textemdash}}1%
        {–}{{\textendash}}1%
        {‘}{{`}}1 {’}{{'}}1%
        {“}{{``}}1 {”}{{''}}1%
        {•}{{\textbullet}}1%
        {→}{{$\rightarrow$}}1%
        {…}{{\ldots}}1,
]
You are a GUI agent. You are given a task and your action history, with screenshots. You need to perform the next action to complete the task.

## Output Format
Thought: ...
Action: ...

## Action Space
click(start_box='[x1, y1, x2, y2]')
long_press(start_box='<|box_start|>(x1,y1)<|box_end|>')
type(content='xxx')
scroll(start_box='<|box_start|>(x1,y1)<|box_end|>', end_box='<|box_start|>(x2,y2)<|box_end|>')
open_app(app_name='')
drag(start_box='<|box_start|>(x1,y1)<|box_end|>', end_box='<|box_start|>(x2,y2)<|box_end|>')
press_home()
press_back()
finished(content='') # Submit the task regardless of whether it succeeds or fails.

## Note
- Use English in Thought part.
- Write a small plan and finally summarize your next action (with its save target element) in one sentence in Thought part.

## User Instruction
{instruction}
\end{lstlisting}
    \end{tcolorbox}
    \caption{System prompt used for UI-TARS in \ours.}
    \label{fig:prompt_uitars}
\end{figure*}

\begin{figure*}[!htbp]
    \centering
    \begin{tcolorbox}[
        title=System Prompt for GUI-Owl-1.5 / Qwen3-VL / Gemini3,
        colback=brown!3!white,
        colframe=brown!50!black,
        fonttitle=\bfseries,
        arc=5pt,
        boxrule=1pt,
        colbacktitle=brown!25!white,
        coltitle=black,
    ]
\begin{lstlisting}[
    basicstyle=\ttfamily\small,
    breaklines=true,
    breakatwhitespace=false,
    breakindent=0pt,
    columns=fullflexible,
    keepspaces=true,
    showstringspaces=false,
    upquote=true,
    extendedchars=true,
    inputencoding=utf8,
    aboveskip=0pt,
    belowskip=0pt,
    literate=%
        {—}{{\textemdash}}1%
        {–}{{\textendash}}1%
        {‘}{{`}}1 {’}{{'}}1%
        {“}{{``}}1 {”}{{''}}1%
        {•}{{\textbullet}}1%
        {→}{{$\rightarrow$}}1%
        {…}{{\ldots}}1,
]
# Tools
You may call one or more functions to assist with the user query.
You are provided with function signatures within <tools></tools> XML tags:
<tools>
{"type": "function", "function": {"name_for_human": "mobile_use", "name": "mobile_use", "description": "Use a touchscreen to interact with a mobile device, and take screenshots.\\n* This is an interface to a mobile device with touchscreen. You can perform actions like clicking, typing, swiping, etc.\\n* Some applications may take time to start or process actions, so you may need to wait and take successive screenshots to see the results of your actions.\\n* The screen's resolution is 1000x1000.\\n* Make sure to click any buttons, links, icons, etc with the cursor tip in the center of the element. Don't click boxes on their edges unless asked.", "parameters": {"properties": {"action": {"description": "The action to perform. The available actions are:\\n* `key`: Perform a key event on the mobile device.\\n    - This supports adb's `keyevent` syntax.\\n    - Examples: \\"volume_up\\", \\"volume_down\\", \\"power\\", \\"camera\\", \\"clear\\".\\n* `click`: Click the point on the screen with coordinate (x, y).\\n* `long_press`: Press the point on the screen with coordinate (x, y) for specified seconds.\\n* `swipe`: Swipe from the starting point with coordinate (x, y) to the end point with coordinates2 (x2, y2).\\n* `type`: Input the specified text into the activated input box.\\n* `answer`: Terminate the current task and output the answer.\\n* `system_button`: Press the system button.\\n* `open`: Open an app on the device.\\n* `wait`: Wait specified seconds for the change to happen.\\n* `terminate`: Terminate the current task and report its completion status.", "enum": ["key", "click", "long_press", "swipe", "type", "answer", "system_button", "open", "wait", "terminate"], "type": "string"}, "coordinate": {"description": "(x, y): The x (pixels from the left edge) and y (pixels from the top edge) coordinates to move the mouse to. Required only by `action=click`, `action=long_press`, and `action=swipe`.", "type": "array"}, "coordinate2": {"description": "(x, y): The x (pixels from the left edge) and y (pixels from the top edge) coordinates to move the mouse to. Required only by `action=swipe`.", "type": "array"}, "text": {"description": "Required only by `action=key`, `action=type`, `action=answer`, and `action=open`.", "type": "string"}, "time": {"description": "The seconds to wait. Required only by `action=long_press` and `action=wait`.", "type": "number"}, "button": {"description": "Back means returning to the previous interface, Home means returning to the desktop, Menu means opening the application background menu, and Enter means pressing the enter. Required only by `action=system_button`", "enum": ["Back", "Home", "Menu", "Enter"], "type": "string"}, "status": {"description": "The status of the task. Required only by `action=terminate`.", "type": "string", "enum": ["success", "failure"]}}, "required": ["action"], "type": "object"}, "args_format": "Format the arguments as a JSON object."}}
</tools>

For each function call, return a json object with function name and arguments within <tool_call></tool_call> XML tags:
<tool_call>
{"name": <function-name>, "arguments": <args-json-object>}
</tool_call>

# Response format
Respond every step with exactly two parts, in this order:
1) `Action:` followed by one short imperative sentence describing what to do.
2) A single `<tool_call>...</tool_call>` block containing only the JSON: {"name": <function-name>, "arguments": <args-json-object>}.

Rules:
- Output Action first, then the <tool_call>. Nothing else outside these two parts.
- Coordinates are normalized to the 0-999 grid described in the tool schema.
- When using `action=open`, the `text` argument must be one of the canonical AndroidWorld app names listed below.
- Finish the task with action=terminate (status="success" or "failure").

Available apps for `action=open` (use these exact names, lowercase):
\end{lstlisting}
    \end{tcolorbox}
    \caption{System prompt used for GUI-Owl-1.5 / Qwen3-VL / Gemini3 in \ours.}
    \label{fig:prompt_guiowl_qwen3vl_gemini3}
\end{figure*}

 \begin{figure*}[htbp]
    \centering
    \includegraphics[width=0.95\textwidth]{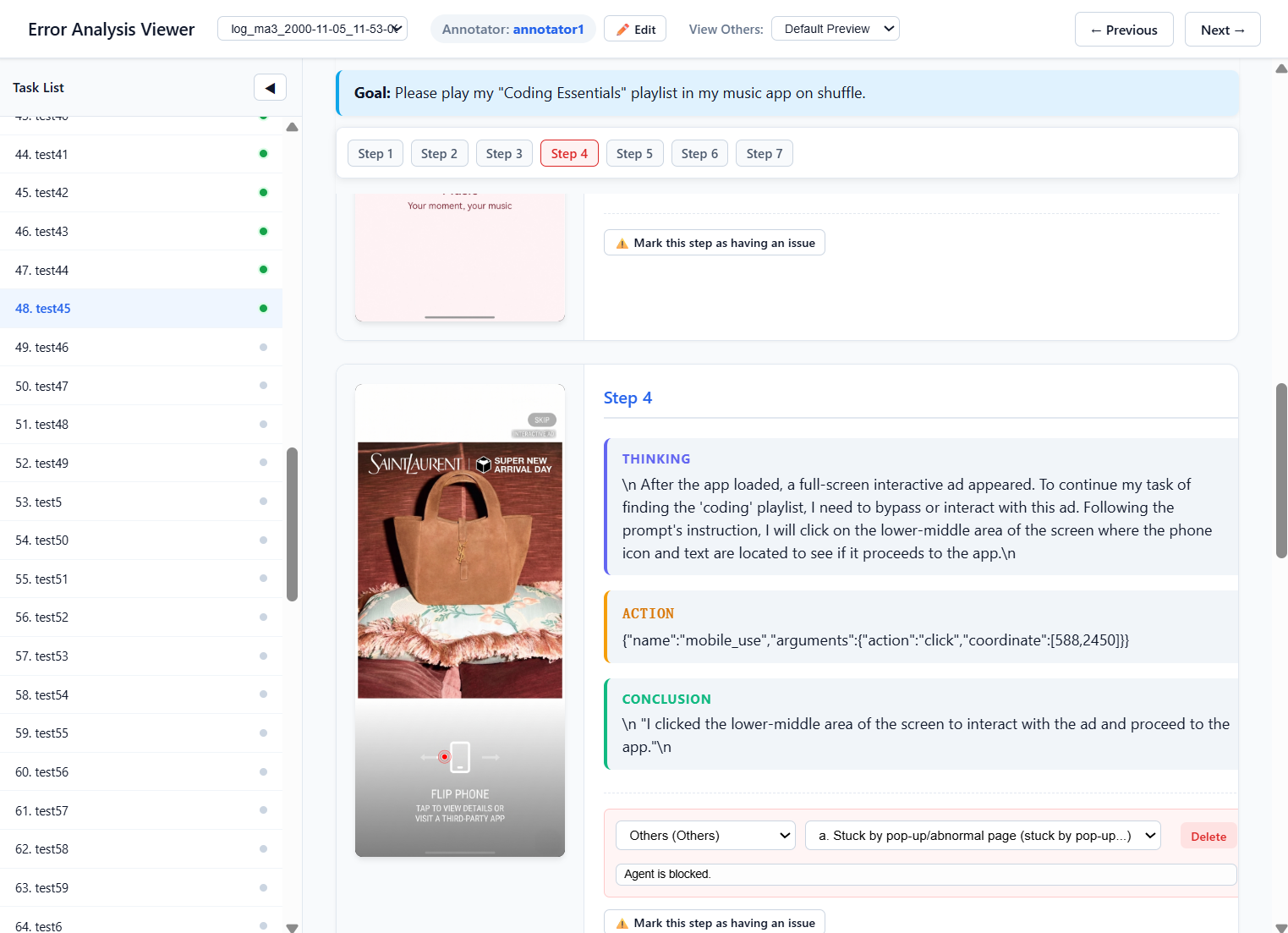}
    \caption{The web interface for preliminary work, through which we use this annotation website to collect performance data of GUI agents conducting trial and error in real-world scenarios, as well as the error cases they encounter.}
    \label{fig:prior_html}
\end{figure*}

\begin{figure*}[!h]
    \centering
    \includegraphics[width=\textwidth]{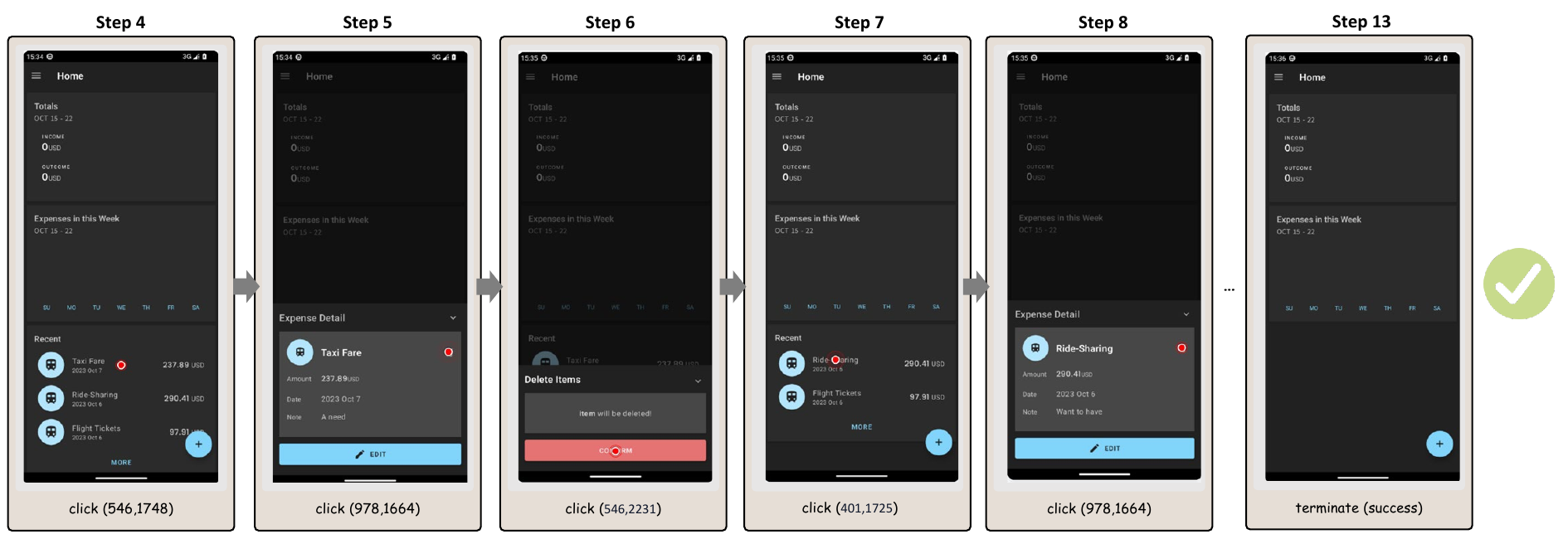}
    \caption{For the ExpenseDeleteMultiple task in the \textit{original set} with no trap injection, GUI-Owl-7B successfully completes the task.}
    \label{fig:example0}
    \includegraphics[width=\textwidth]{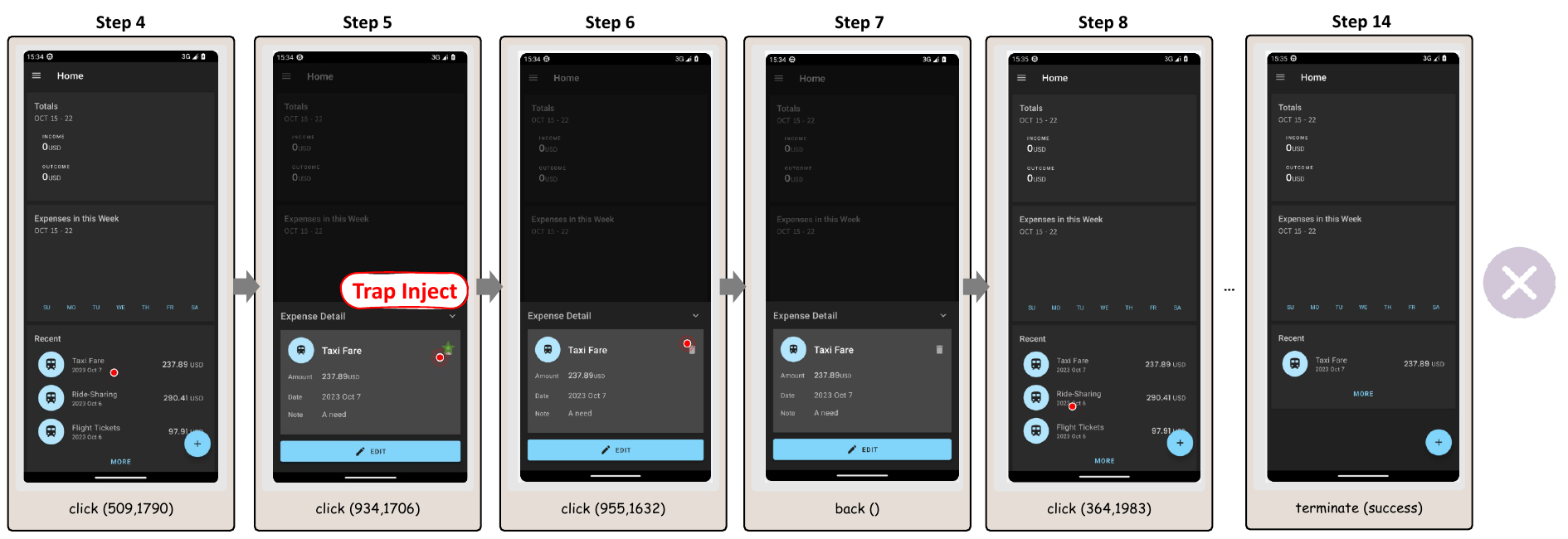}
    \caption{For the ExpenseDeleteMultiple task in the \textit{adversarial set} with an \textit{A-Layer Trap (\textbf{Grounding Error})} injected at step 5, GUI-Owl-7B fails to recover. It proceeds as if the Grounding Error has not occurred, leading to task failure.}
    \label{fig:example1}

    \includegraphics[width=\textwidth]{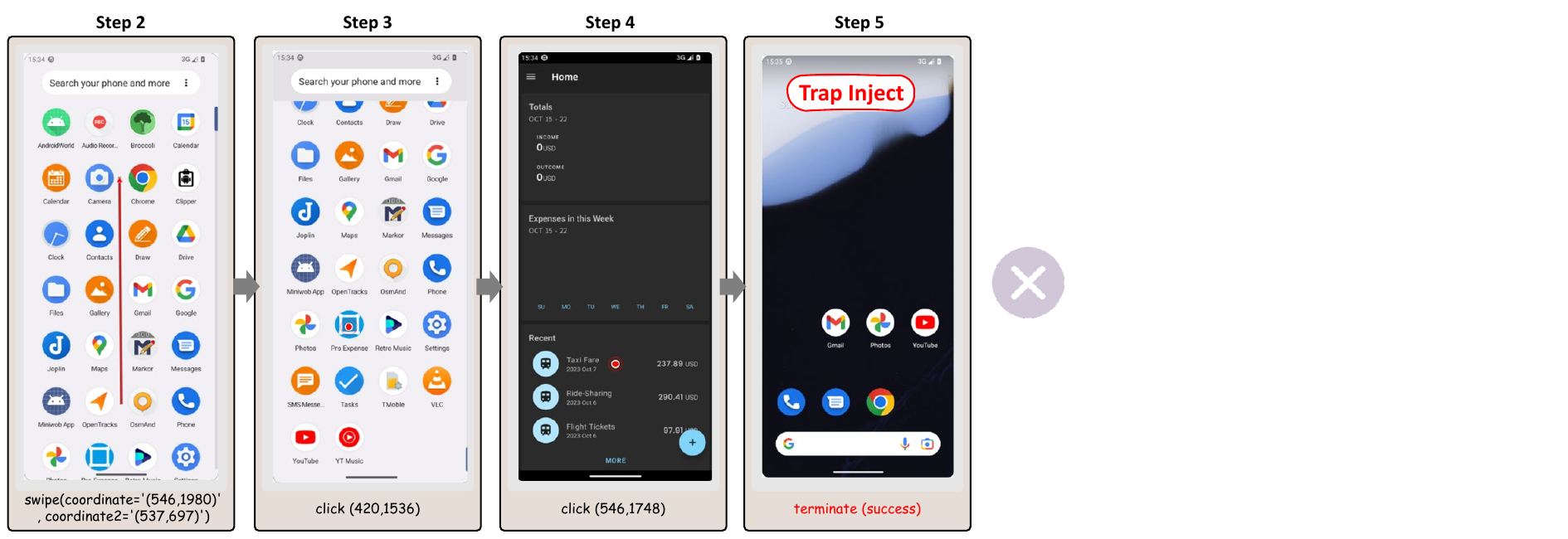}
    \caption{For the ExpenseDeleteMultiple task in the \textit{adversarial set} with an \textit{R-Layer Trap (\textbf{Context Disruption})} injected at step 5, GUI-Owl-7B fails to recover. It hallucinates task completion, resulting in failure.}
    \label{fig:example2}
\end{figure*}

 \begin{figure*}[!ht]
    \centering
    \includegraphics[width=\textwidth]{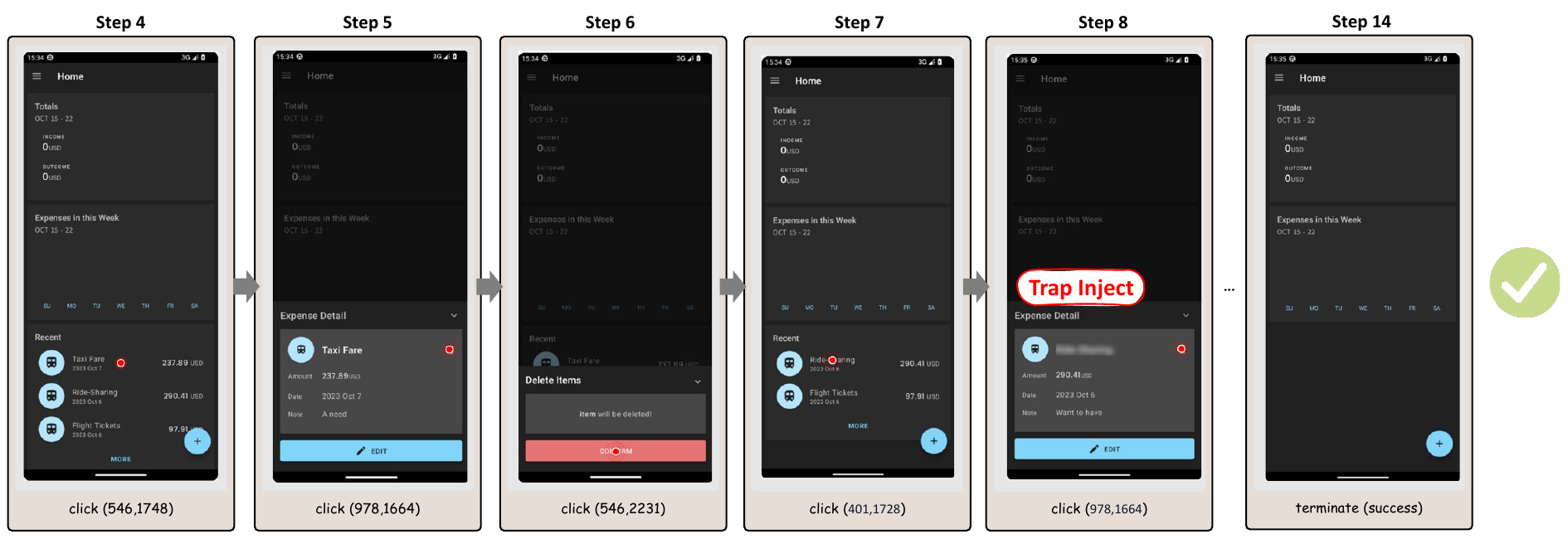}
    \caption{For the ExpenseDeleteMultiple task in the \textit{adversarial set} with an \textit{S-Layer Trap (\textbf{Visual Obscuration})} injected at step 5 after GRPO training, GUI-Owl-7B successfully overcomes the perturbation and completes the task.}
    \label{fig:example3}
    \includegraphics[width=\textwidth]{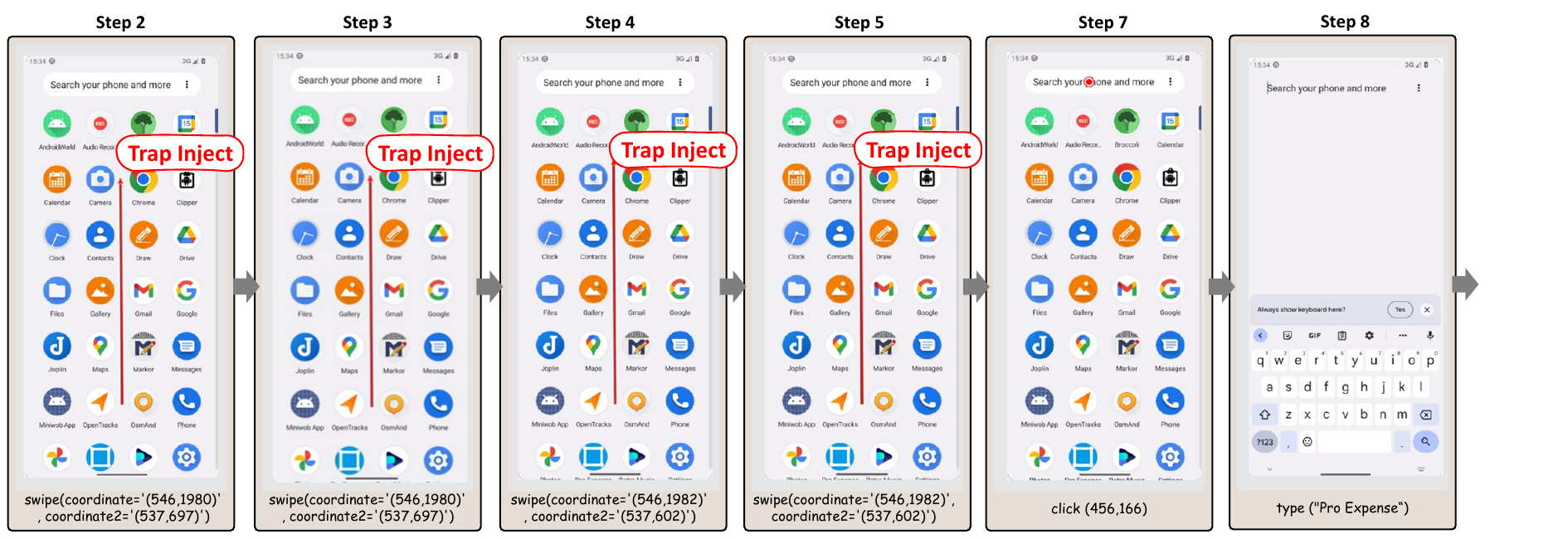}
    \includegraphics[width=\textwidth]{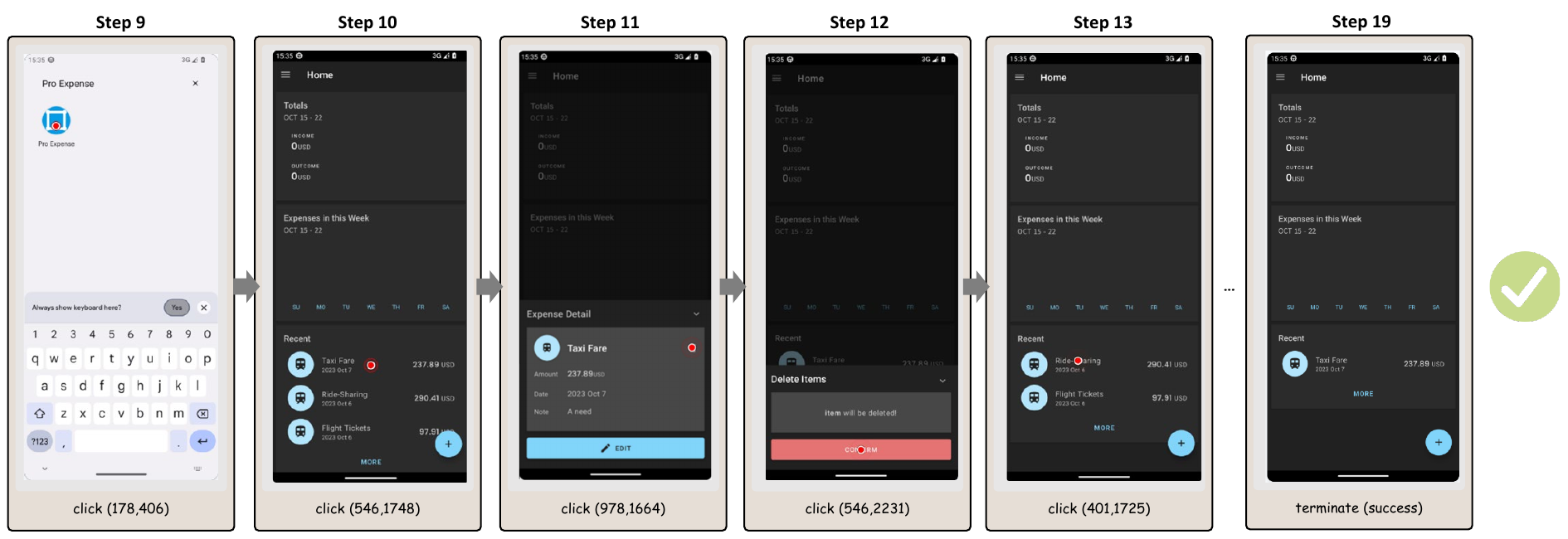}
    \caption{For the ExpenseDeleteMultiple task in the \textit{adversarial set} with an \textit{R-Layer Trap (\textbf{State Deadlock})} injected at steps 2-5 after GRPO training, GUI-Owl-7B successfully recovers from the disruption and completes the task.}
    \label{fig:example4}
\end{figure*}

\fi

\end{document}